\documentclass[10pt,twocolumn,letterpaper]{article}

\usepackage{cvpr}              % To produce the CAMERA-READY version
\usepackage{graphicx}   % for \resizebox
\usepackage{booktabs}   % for \toprule, \midrule, \bottomrule
\usepackage{makecell}   % for \makecell
\usepackage{multirow}

\usepackage{amsmath,amssymb}
\usepackage{algorithm}          % 提供 \begin{algorithm} 浮动体
\usepackage{algpseudocode}      % 提供 \begin{algorithmic} 环境（推荐）
\algrenewcommand\algorithmicrequire{\textbf{Input:}}
\algrenewcommand\algorithmicensure{\textbf{Output:}}

\usepackage{tabularx,array}

\usepackage{capt-of} % in the preamble
\usepackage[dvipsnames,svgnames,x11names]{xcolor}
\usepackage{soul}

\definecolor{MismatchPink}{HTML}{FFC2D1} % 
\definecolor{MismatchRose}{HTML}{F4A7B9} % 
\definecolor{MatchGreen}{HTML}{CDECCF}   % 
\definecolor{MatchBlue}{HTML}{CFE8FF}    % 
\definecolor{oraclecolor}{HTML}{C74634}
\definecolor{ZeroReward}{HTML}{F59E0B}

\newcommand{\mismatching}[1]{%
  {\sethlcolor{MismatchPink}\hl{#1}}%
}
\newcommand{\matching}[1]{%
  {\sethlcolor{MatchGreen}\hl{#1}}%
}
\definecolor{cvprblue}{rgb}{0.21,0.49,0.74}
\usepackage[pagebackref,breaklinks,colorlinks,allcolors=cvprblue]{hyperref}

\def\paperID{8320} % *** Enter the Paper ID here
\def\confName{CVPR}
\def\confYear{2025}

\title{\textcolor{oraclecolor}{OraPO}: Oracle-educated Reinforcement Learning for \\Data-efficient and Factual Radiology Report Generation}

\author{Zhuoxiao Chen$^{1,2\dagger}$, Hongyang Yu$^{1\star}$, Ying Xu$^{1}$, Yadan Luo$^{2}$, Long Duong$^1$, Yuan-Fang Li$^{1\star}$ \\
$^1$Oracle Health \& AI, 
$^2$The University of Queensland\\
\tt\small\{ivan.chen, hongyang.yu, ying.x.xu, long.duong, yuanfang.li\}@oracle.com \\ \tt\small\{zhuoxiao.chen, y.luo\}@uq.edu.au
\vspace{-1ex}
}

\begin{document}
% \maketitle
% \let\thefootnote\relax\footnotetext{$^\dagger$ Intern of Oracle, $^{\star}$ Corresponding author.}

\twocolumn[{%
\renewcommand\twocolumn[1][]{#1}

\maketitle

\begin{center}
\includegraphics[width=1\linewidth]{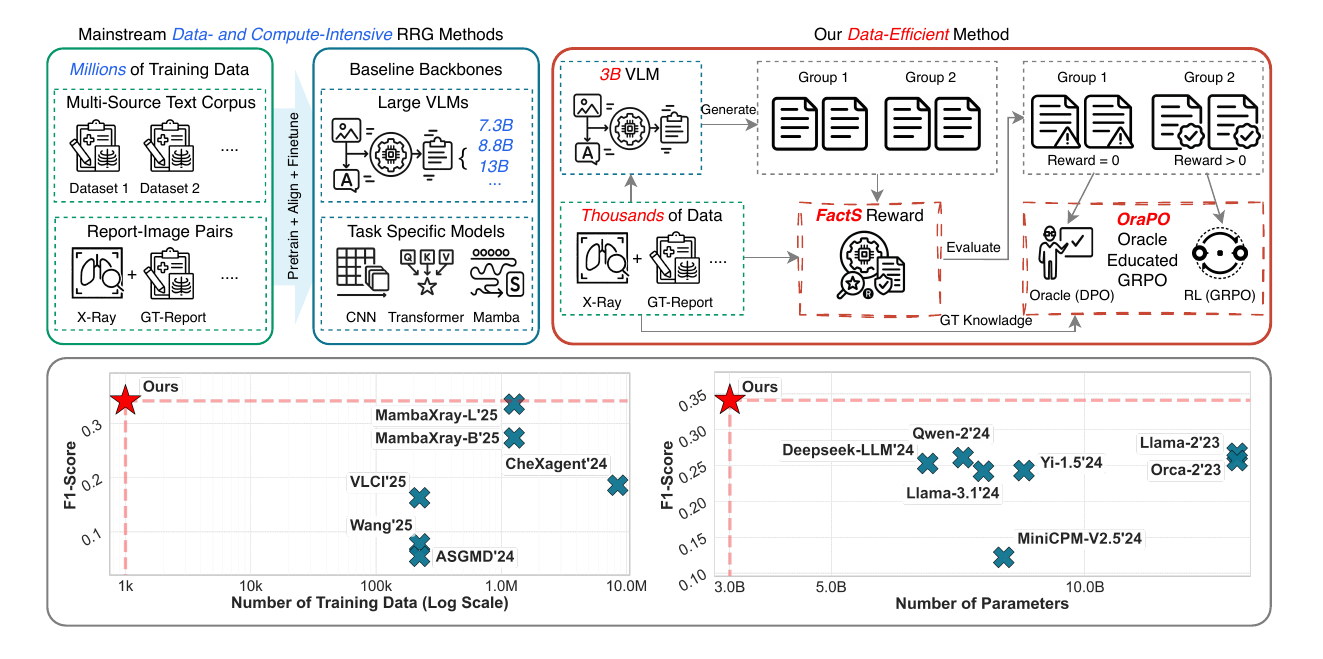}
  \captionof{figure}{
  Illustration of mainstream data/compute-intensive pipelines (\emph{upper-left}) versus our data-efficient pipeline (\emph{upper-right}). \emph{Bottom:} on CheXpert Plus \cite{DBLP:journals/corr/abs-2405-19538}, our method achieves the \textbf{SOTA performance for RRG} with \textbf{less than 0.1\%} of the training samples (vs. 1.27 M) used by best-performing baselines and a \textbf{much smaller model}, demonstrating strong performance under tight data and compute budgets.}
  \label{fig:efficient_setting}
\end{center}
}]

% --- author footnote under the title, full width ---
\begingroup
\renewcommand\thefootnote{}% no mark
% \footnotetext{$^\dagger$ Work done during an internship at Oracle Health \& AI. $^{\star}$ Joint senior authors.}
\addtocounter{footnote}{-1}% keep numbering correct
\endgroup
% ---------------------------------------------------

\begin{abstract}
Radiology report generation (RRG) aims to automatically produce clinically faithful reports from chest X-ray images. Prevailing work typically follows a \textbf{scale-driven paradigm}, by multi-stage training over large paired corpora and oversized backbones, making pipelines highly data- and compute-intensive. In this paper, we propose Oracle-educated GRPO (\textcolor{oraclecolor}{\textbf{OraPO}}) with a FactScore-based reward (\textcolor{oraclecolor}{\textbf{FactS}}) to tackle the RRG task under constrained budgets.  OraPO enables \textit{single-stage}, \textit{RL-only} training by converting failed GRPO explorations on rare or difficult studies into direct preference supervision via a lightweight oracle step. FactS grounds learning in diagnostic evidence by extracting atomic clinical facts and checking entailment against ground-truth labels, yielding dense, interpretable sentence-level rewards. Together, OraPO and FactS create a compact and powerful framework that significantly improves learning efficiency on clinically challenging cases, setting the \textbf{new SOTA} performance on the CheXpert Plus dataset (0.341 in F1) and MIMIC-CXR dataset (0.357 in F1) with \textbf{2--3 orders of magnitude less training data} using a \textbf{small} base vision-language model on modest hardware.

% Together, OraPO and FactS form a data- and compute-efficient framework that significantly improves performance on clinically challenging cases, achieving state-of-the-art results on CheXpert Plus (0.341 F1) with with only 1k training studies (2–3 orders of magnitude fewer training samples than prior work) using a small-sized VLM on modest hardware.

%Trained with only 1k samples (\textit{i.e.}\ \textbf{1{,}000$\times$} less data than the best-performing baseline) on CheXpert Plus datasets, our approach achieves the \textbf{state-of-the-art clinical effectiveness (0.341 in F1)}, outperforms the best-performing baseline by \textbf{161\%} in recall, with a far smaller model (3B), delivering a practical, data- and compute-efficient solution to RRG’s core obstacles.
 \end{abstract}
    
\section{Introduction}
\label{sec:intro}
% \noindent \textbf{[Paragraph 1, Old Version]} 
% Radiology report generation (RRG) from chest X-rays aims to turn imaging evidence into a clinically usable narrative. This is a complex multimodal task that couples disease prediction with free-text report generation, linking detected findings (\textit{e.g.} edema, consolidation, atelectasis, pleural effusion, and pneumothorax) to clear sentences that describe their presence, details, location, and severity. Automating this task matters in practice: the large imaging volumes and backlogs place serious strains on healthcare services and delay patient care, while radiologist staffing remains insufficient. For example, a 29\% radiology consultant shortfall is reported alongside persistent vacancies in England \cite{Nightingale2024BJROpen}. In the USA, there were more than 14,000 job postings for radiologists in 2022, while only 1,150 residents graduate each year \cite{doi:10.2214/AJR.24.30920}. In this context, AI-assisted drafting has begun to show measurable efficiency gains. A recent study \cite{DBLP:journals/corr/abs-2412-12042} reported that using draft reports generated by large language models (LLMs) reduced reporting time from 573s to 435s on average (\textit{i.e.}\ around 24\% faster). Given its practical importance, automated radiology report generation has rapidly become an active research area aimed at easing radiologists' workload and streamlining workflows in hospitals and clinics.

% \noindent
Radiology report generation (RRG) from chest X-rays aims to turn imaging evidence into a clinically usable narrative. This is a complex multimodal task that couples disease prediction with free-text report generation, linking detected visual findings (\textit{e.g.} edema, pleural effusion, and pneumothorax) to clear sentences that describe their presence, details, location, and severity. Automating this task matters in practice: the large imaging volumes and backlogs place serious strains on healthcare services and delay patient care, while radiologist staffing remains insufficient. For example, England reports a 29\% radiology consultant shortfall alongside sustained vacancies \cite{Nightingale2024BJROpen}, while in the United States there are about 14{,}000 openings but only 1{,}150 resident graduates annually \cite{doi:10.2214/AJR.24.30920}. In this context, AI-assisted drafting has begun to show measurable efficiency gains. A recent study~\cite{DBLP:journals/corr/abs-2412-12042} reported that using draft reports generated by large language models (LLMs) reduced reporting time by 24\%. Given its practical importance, automated radiology report generation has rapidly become an active research area aimed at easing radiologists' workload and streamlining workflows in hospitals and clinics.

Mainstream RRG methods typically rely on supervised fine-tuning (SFT) via multi-stage training on large corpora and/or scaling up backbone size (Fig.~\ref{fig:efficient_setting}, upper-left). The multi-stage training paradigm usually follows three phases: domain-specific pre-training, image–text alignment, and task-specific fine-tuning, where the first two are usually trained on combinations of large-scale, multi-source datasets such as medical textbooks, real radiology reports including but not limited to MIMIC-CXR \cite{DBLP:journals/corr/abs-1901-07042}, IU X-ray \cite{DBLP:journals/jamia/Demner-FushmanK16} and CheXpert Plus \cite{DBLP:journals/corr/abs-2405-19538}, often for dozens of epochs \cite{DBLP:conf/cvpr/WangWLMW0025,DBLP:conf/cvpr/WangLWZ23}. Notably, both CheXpert Plus and MIMIC-CXR include training splits \textbf{exceeding 200K} image-report pairs. In addition, there is a recent trend of adopting large vision-language models (VLMs) with $>$13B parameters for SFT, demanding meticulous data curation and substantial GPU budgets \cite{DBLP:conf/nips/LiWZULYNPG23,DBLP:conf/miccai/KimWAGW24}.

Recently, \emph{Group Relative Policy Optimisation} (GRPO) has emerged as a strong and compute-efficient RL-based learning algorithm for many hard problems with verifiable rewards. Its efficiency stems from two fronts: (i) lower memory/compute by dispensing with the trainable value critic and instead computing group-normalised advantages from the relative rewards of multiple samples \cite{DBLP:journals/corr/abs-2505-22257, DBLP:journals/corr/abs-2504-20834}; and (ii) a simplified training recipe that enables ``RL-only'' regimes (\textit{i.e.}, without costly SFT/alignment), avoiding massive pre-training/alignment corpora \cite{DBLP:journals/corr/abs-2402-03300, DBLP:conf/icdm/ZhangXQLW25, DBLP:journals/corr/abs-2501-12948,DBLP:journals/corr/abs-2504-11468}. 

\begin{figure}[t]
  \centering
    \includegraphics[width=1\linewidth]{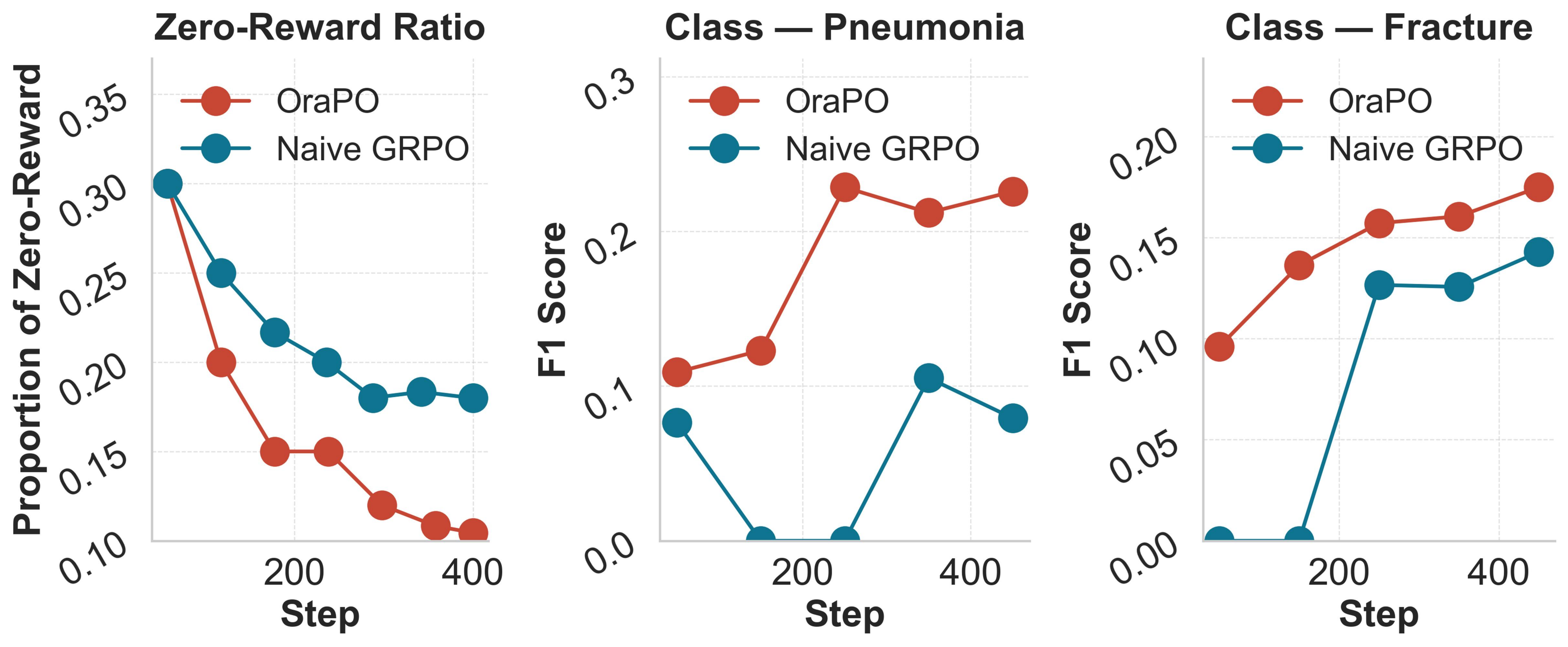}
      \captionof{figure}{
      \emph{Left:} Cumulative proportion of \emph{zero-reward} batches (reward batch mean = 0) vs. training step on CheXpert Plus \cite{DBLP:conf/cvpr/WangWLMW0025}. \textcolor{oraclecolor}{\textbf{OraPO}} suppresses zero-reward frequency faster than na\"ive GRPO. \emph{Centre/Right:} Class-level F1 on the CheXpert Plus validation set \cite{DBLP:conf/cvpr/WangWLMW0025} across checkpoints for two\textit{ clinically challenging} and \textit{rare} classes: \textbf{Pneumonia} (2.70\%) and \textbf{Fracture} (4.05\%). \textcolor{oraclecolor}{\textbf{OraPO}} learns earlier and maintains higher F1 than na\"ive GRPO.}
  \label{fig:zero_reward_illustration}
\end{figure}

However, applying GRPO to the RRG task reveals two challenges. \underline{Challenge (i)}: we observe that vanilla GRPO often produces all-zero reward groups early in training (\textit{e.g.} 30\% in the first 50 steps, as shown in Fig.~\ref{fig:zero_reward_illustration} left), causing vanishing gradients and wasted rollouts. Recent variants attempt to fix this by resampling until a non-zero reward appears \cite{DBLP:journals/corr/abs-2503-14476} or by enlarging the group size \cite{DBLP:journals/corr/abs-2508-09726}, both of which are still suboptimal due to \emph{increased cost} of more rollouts. Another line interleaves SFT with RL to maintain gradient flow, yet still leaves low-quality rollouts unused, wasting costly sampling \cite{DBLP:journals/corr/abs-2506-07527,DBLP:journals/corr/abs-2509-6948,DBLP:journals/corr/abs-2508-11408,DBLP:journals/corr/abs-2509-04419,DBLP:journals/corr/abs-2505-16984}. \underline{Challenge (ii):} vanilla GRPO requires readily verifiable scalar rewards. Unlike in multiple-choice QA, maths, or coding, where the reward is typically a binary final-answer check, radiology reports are long-form, multi-fact, and must be consistent across sentences, making reward design both critical and difficult. In practice, RLHF-style RRG work often uses reference-overlap metrics that mainly capture \emph{fluency/surface similarity} (\textit{e.g.} BLEU/CIDEr \cite{DBLP:conf/nips/LiLHX18,DBLP:conf/bionlp/NicolsonLDNK24}) or a \emph{report-level clinical metric} \cite{DBLP:journals/corr/abs-2403-06728,DBLP:conf/naacl/MiuraZTLJ21,DBLP:journals/patterns/YuEKPTRFLANLVR23}. These proxies under-penalise sentence-level factual errors and cross-sentence contradictions, yielding fluent yet clinically misleading narratives (missed positives, unsupported claims). Hence, a reward must be able to assess factual correctness of each statement of a radiology report in an efficient manner.

In this paper, we propose \underline{Ora}cle-educated Group Relative \underline{P}olicy \underline{O}ptimisation: \textcolor{oraclecolor}{\textbf{OraPO}}, a novel RL-based learning algorithm that tackles RRG's challenges in a data- and compute-efficient manner. OraPO addresses \underline{Challenge (i)} of GRPO by dynamically injecting oracle supervision (\textit{i.e.}\ ground-truth) for groups with \textit{all-zero} rewards and teaching the policy to avoid repeating low-quality generations. Specifically, When a sampled group yields \emph{all-zero} rewards (group 1 in Fig.~\ref{fig:efficient_setting}, upper-right), OraPO triggers a lightweight Direct Preference Optimisation (DPO) update that directly prefers the ground-truth report over the zero-reward rollouts, effectively reusing those failed rollouts as  \emph{ready-made negative} examples, with no extra compute or annotation needed. This converts failed exploration in difficult cases (hard or low-prevalence studies) into useful gradients, moving the policy away from low-quality modes and toward the ground-truth. We also introduce a simple adaptive mixing weight to increase DPO's influence when zero-reward groups are frequent, which gradually tapers once GRPO provides informative advantages, thereby allowing OraPO to salvage otherwise wasted batches, stabilise training, and accelerate convergence.
.

% [HY Version]
% A common theme of the above pioneering work is the use of Group Relative Policy Optimisation (GRPO) with verifiable reward design. This poses two major challenges for extending this line of work to solving the much harder radiology report generation task. First, radiology report generation is a highly specialized and specific task that requires domain knowledge often missing in popular pretrained backbone. Due to the lack of such knowledge, at the early stage of training, a lot of compute is wasted on generating rollout groups that all yield 0 reward, stalling training and prevent the model from learning. Applying GRPO directly on these pretrained backbones will unlikely yield noticeable improvement. A recent comprehensive study \cite{yue2025doesreinforcementlearningreally} systematically verified the capacity of RL learning is limited by the knowledge capacity of the base model. Secondly, the na\"ive version of GRPO requires directly and deterministically verifiable reward which is challenging to obtain from the radiology report generation perspective. There needs to be a more flexible and medically grounded way of verifying the correctness of a radiologist report.

Moreover, to provide OraPO with clinically faithful supervision, we introduce a \textcolor{oraclecolor}{\textbf{FactS}}core-based reward \cite{DBLP:conf/emnlp/MinKLLYKIZH23} designed to address \underline{Challenge (ii)}. We treat the generated report as its own rationale, extract \emph{atomic clinical facts}, and test their \emph{entailment} against the ground-truth label set. Facts that support a label contribute positively; unsupported or contradictory labels incur penalties, yielding a dense, interpretable per-label reward. This stands in sharp contrast to the original GRPO reward, which assigns reward solely to the final answer while treating the reasoning trace largely as a byproduct of generation, without ensuring it is factually grounded or consistently aligned with the answer. Our reward design ensures the generated report (serving in lieu of reasoning trace) not only produces a correct cognitive reasoning trace similar to that of an experienced radiologist but also yields correct diagnostic findings, which is crucial in domains such as healthcare in which safety is of paramount importance.

Extensive experiments are conducted on CheXpert Plus \cite{DBLP:journals/corr/abs-2405-19538} and MIMIC-CXR \cite{DBLP:journals/corr/abs-2405-19538} datasets, where OraPO sets the new SOTA performance for RRG, surpassing the previous best model with F1 score and notably achieving \textbf{160.8\% improvement in recall}. This higher recall is clinically meaningful. In automated RRG tasks, recall is often prioritised over precision, as false negatives (missed abnormalities) can delay or prevent timely treatment, whereas false positives generally result only in additional clinical review \cite{DBLP:conf/naacl/MiuraZTLJ21, seyyed2021underdiagnosis, rao2024rexerr, noguchi2023analysis}. Previous studies \cite{lakhani2012automated, DBLP:journals/patterns/YuEKPTRFLANLVR23} emphasise maintaining high recall to avoid critical finding omissions, recognising that such oversights carry more serious clinical consequences than reduced precision. The proposed method achieves this clinical effectiveness while being exceptionally data-efficient, using \textbf{2–3 orders of magnitude less training data} (\textit{i.e.} trained on only \textbf{1K} samples compared to the current best baseline MambaXray-L’s \textbf{1.27M} samples and other baselines’ minimum of 223K samples). At the same time, OraPO is compute-efficient: it is based on Qwen2.5-VL-\textbf{3B}, a small-sized VLM on modest hardware (\textbf{4× A10 GPUs}).

In summary, our contributions are \textbf{threefold}:
\begin{enumerate}
\item We present \textbf{OraPO}, an Oracle (DPO) educated GRPO that turns failed explorations into direct preference supervision significantly improving data- and compute-efficiency. To the best of our knowledge, this is the \textbf{first} work to integrate direct preference learning with GRPO-based RL for efficient training. 
\item We design a \textbf{FactS Reward} for RL, performing fact-level entailment against ground-truth labels, yielding dense, interpretable feedback that aligns reports with diagnostic facts while avoiding long reasoning traces.
\item We show that OraPO achieves new SOTA performance for RRG, surpassing the previous best model (MambaXray-L) in clinical effectiveness F1 and recall while using \textbf{2--3 orders of magnitude less training data}.

% \item OraPO sets the new SOTA performance for RRG, surpassing the previous best model in terms of clinical effectiveness F1 score. This is achieved while being exceptionally data-efficient, with \textbf{2--3 orders of magnitude less training data}. OraPO is trained on only \textbf{1K} samples. In contrast, the current best model MambaXray-L is trained on \textbf{1.27M} samples and all the other models are trained on at least 223K samples. At the same time, OraPO is also compute-efficient: it is based on QWen2.5-VL-\textbf{3B}, a moderately-sized VLM on modest hardware (\textbf{4x A10 GPUs}).

\end{enumerate}

% \begin{figure}[t]
%   \centering
%     \includegraphics[width=1\linewidth]{figures/labeler_compare.pdf}
%       \captionof{figure}{
%       Alignment with human gold labels on the CheXpert Plus validation set. The plot compares three labelers: CheXbert, CheXpert, and an LLM-based entailment labeler (GPT\mbox{-}4.1 \cite{DBLP:journals/corr/abs-2303-08774}), using macro-averaged Precision, Recall, and F1 across 14 pathologies. Overall, the GPT\mbox{-}4.1 labeler shows the highest alignment (best P/R/F1), CheXpert \cite{DBLP:journals/corr/abs-2405-19538} ranks second, and CheXbert \cite{DBLP:journals/corr/abs-2004-09167} trails behind.}
%   \label{fig:zero_reward_illustration}
% \end{figure}

\section{Related Work}
\noindent\textbf{Automatic Radiology Report Generation (RRG)} aims to turn imaging studies (\textit{e.g.}, chest X-rays) into clinically usable free-text reports. Early approaches adopted seq2seq networks (\textit{e.g.}, CNN–RNN and Transformer) \cite{DBLP:conf/acl/XingXJ18, DBLP:conf/emnlp/ChenSCW20, DBLP:conf/nips/LimC0B0HL24, DBLP:journals/tmlr/YuY24, DBLP:journals/bioinformatics/YuY22} for training on large paired corpora such as MIMIC-CXR \cite{DBLP:journals/corr/abs-1901-07042}. Recent work advances three fronts: (i) \emph{knowledge- and rule-guided generation}, which injects medical priors to curb hallucination \cite{DBLP:conf/cvpr/BuLYD24} and performs disease-aware image–text alignment for trustworthy text \cite{DBLP:conf/cvpr/ParkHSSOK25, DBLP:conf/emnlp/JiangZZJYL24}; (ii) \emph{leveraging prior studies and multi-view images} to improve coverage and consistency \cite{DBLP:conf/miccai/HamamciEM24, DBLP:conf/cvpr/LiuMK0XJM25, DBLP:journals/pami/ChenLWWH25}; and (iii) \emph{scaling/standardizing with pre-training}, releasing multi-stage pre-training recipes \cite{DBLP:conf/cvpr/WangWLMW0025, DBLP:conf/miccai/HamamciEM24, DBLP:conf/iccv/ChenL0BH23, DBLP:conf/iclr/ChenMB0HL25, DBLP:journals/corr/abs-2510-08668}. In parallel, LLM-driven RRG explores instruction-tuned or preference-aligned training \cite{DBLP:conf/aaai/LiuTCS024, DBLP:conf/icml/TuDC0ZGL25, DBLP:conf/nips/LiWZULYNPG23}. Despite progress, these approaches often rely on large paired corpora and/or costly computation resources, leading to deficiencies in data efficiency and practical scalability.

\noindent\textbf{Group Relative Policy Optimisation (GRPO)} is a critic-free, reinforcement learning \emph{objective} for generative LLMs, introduced in \emph{DeepSeekMath} \cite{DBLP:journals/corr/abs-2402-03300}. It samples multiple completions per prompt, computes group-normalised advantages from relative rewards, and updates the policy with clipped ratios to a frozen reference; removing the critic reduces memory/compute versus PPO while preserving stability \cite{DBLP:journals/corr/abs-2402-03300,guo2025deepseeknature, DBLP:conf/cikm/ZhangQL024, DBLP:journals/corr/abs-2505-22257,DBLP:journals/corr/abs-2504-20834}. Recent variants mainly target optimisation stability: DR.GRPO \cite{DBLP:journals/corr/abs-2503-20783} mitigates length bias and skewed group rewards; CISPO \cite{DBLP:journals/corr/abs-2506-13585} clips importance-sampling weights; GSPO \cite{DBLP:journals/corr/abs-2507-18071} adopts sequence-level clipping for MoE; DAPO \cite{DBLP:journals/corr/abs-2503-14476} resamples until a non-zero reward appears. However, these methods were not designed to improve data efficiency: most target optimisation stability while leaving data volume and training epochs essentially unchanged, and some of them often increase sampling/compute \cite{DBLP:journals/corr/abs-2503-14476, DBLP:journals/corr/abs-2506-22950, DBLP:conf/mm/ChenWL0H24, DBLP:journals/corr/abs-2508-09726}. 

\noindent\textbf{Direct Preference Optimisation (DPO)} trains a policy from \emph{preference pairs} to prefer the chosen response over the rejected one \cite{zhong2024dpo}. Recent variants \cite{, yuan2026mitigating} streamline or improve the robustness of this recipe: SimPO \cite{DBLP:conf/nips/0001X024} applies length-normalised mean log-probabilities with a simplified reference; Cal-DPO \cite{DBLP:conf/nips/XiaoYZLH24} calibrates the implicit reward to human scales; ORPO \cite{DBLP:conf/emnlp/HongLT24} drops the reference via an odds-ratio objective integrated into SFT; and KTO \cite{DBLP:journals/corr/abs-2402-01306} learns from unary thumbs-up/down signals instead of pairs.  In our RRG framework, DPO naturally serves as an \emph{oracle step} inside \textbf{OraPO} to convert those failed GRPO rollouts into useful preference updates against the ground-truth, 
leading to data- and compute-efficient learning.

\label{sec:related_work}

\section{Methodology}

In this section, we first introduce the background of Group Relative Policy Optimisation (GRPO) and Direct Preference Optimisation (DPO) (\S\ref{sec:preliminaries}). We then identify a failure mode of GRPO in radiology report generation (RRG), when a pretrained model produces many low-quality groups (\textit{e.g.}, zero rewards) that yield little or no gradient. This observation motivates our proposed {\textbf{OraPO}} algorithm, which efficiently converts such failed explorations into preference supervision (\S\ref{sec:OraPO}). Finally, we design the {\textbf{FactS Reward}}, a fact-level entailment signal that aligns reports with diagnostic labels and provides dense, interpretable feedback for reinforcement learning (\S\ref{sec:facts_reward}).

\subsection{Preliminaries}\label{sec:preliminaries}

\noindent\textbf{Group Relative Policy Optimisation (GRPO).} 
GRPO~\cite{DBLP:journals/corr/abs-2402-03300, guo2025deepseeknature} is a computation- and memory-efficient RL objective for aligning generative models (LLMs/VLMs), introduced for reasoning-heavy settings and now used broadly because of its efficient critic-free design and proven effectiveness.
It replaces the value critic with \emph{group-normalized} advantages computed among multiple completions sampled for the same input. Given a study–prompt pair $(x,p)$, the policy $\pi_{\theta}$ draws a group of $K$ sequences $\{\hat y_{j}\}_{j=1}^{K}\sim \pi_{\theta}(\cdot\mid x,p)$ and receives scalar rewards $r_{j}=r(x,\hat y_{j})$.
GRPO forms a group baseline and scale,
\begin{equation}
\begin{aligned}
\bar r=\frac{1}{K}\sum_{j=1}^{K} r_{j},\qquad
\sigma=\sqrt{\frac{1}{K}\sum_{j=1}^{K}(r_{j}-\bar r)^{2}+\varepsilon},
\end{aligned}
\end{equation}
and defines the group-normalised advantage:
\begin{equation}
\begin{aligned}
A_{j}=\frac{r_{j}-\bar r}{\sigma}.
\end{aligned}
\end{equation}
Let the proximal policy optimisation (PPO) ratio be:
\begin{equation}
\begin{aligned}
\rho_{j}=\exp\!\big(\log \pi_{\theta}(\hat y_{j}\mid x,p)-\log \pi_{\theta_{\text{old}}}(\hat y_{j}\mid x,p)\big),
\end{aligned}
\end{equation}
where $\theta_{\text{old}}$ denotes the parameters of the behaviour policy used to sample the current group (\textit{i.e.}, the parameters before the update). Then the GRPO loss over a batch is defined as:
\begin{equation}
\begin{aligned}
\mathcal{L}_{\mathrm{GRPO}}
&= -\,\mathbb{E}\Big[\min\big(\rho_{j}A_{j},\ \mathrm{clip}(\rho_{j},\,1-\epsilon,\,1+\epsilon)\,A_{j}\big)\Big] \\
&\quad + \lambda_{\mathrm{KL}}\ \mathbb{E}\!\left[\mathrm{KL}\!\big(\pi_{\theta}(\cdot\mid x,p)\,\|\,\pi_{\mathrm{ref}}(\cdot\mid x,p)\big)\right],
\end{aligned}
\label{eq:grpo}
\end{equation}
where $\epsilon$ is the clipping threshold, $\lambda_{\mathrm{KL}}\ge 0$ controls regularization toward a frozen reference policy $\pi_{\mathrm{ref}}$.
In practice, GRPO samples $K$ completions per prompt, computes $r_j$, normalizes within-group to obtain $A_j$, and updates $\pi_{\theta}$ with the clipped-ratio objective in~\eqref{eq:grpo}, thereby achieving critic-free policy optimisation.

\noindent\textbf{Direct Preference Optimisation (DPO).} DPO \cite{zhong2024dpo} is a widely used, preference-alignment objective for generative LLMs/VLMs. It updates the policy to prefer \emph{chosen} completions over \emph{rejected} ones relative to a frozen reference model.
Given a preference dataset $\mathcal{P}=\{(x,p, y^{+}, y^{-})\}$, where $y^{+}$ is preferred to $y^{-}$ for the same input–prompt pair $(x,p)$, DPO maximizes the log-probability margin of the preferred sample after subtracting the reference log-probabilities.
Let $\pi_{\theta}$ be the trainable policy and $\pi_{\mathrm{ref}}$ the frozen reference. The expectation below is taken over preference pairs drawn from $\mathcal{P}$. With temperature $\tau>0$, define
\begin{equation}
\Delta^{+}
=
\log \pi_{\theta}(y^{+}\mid x,p)\;-\;\log \pi_{\mathrm{ref}}(y^{+}\mid x,p),
\label{eq:dpo-delta-plus}
\end{equation}
\begin{equation}
\Delta^{-}
=
\log \pi_{\theta}(y^{-}\mid x,p)\;-\;\log \pi_{\mathrm{ref}}(y^{-}\mid x,p),
\label{eq:dpo-delta-minus}
\end{equation}
and the DPO loss
\begin{equation}
\mathcal{L}_{\mathrm{DPO}}
=
-\,\mathbb{E}\Big[
\log \sigma\!\big(\tau\,(\Delta^{+}-\Delta^{-})\big)
\Big],
\label{eq:dpo}
\end{equation}
where $\sigma(\cdot)$ is the logistic sigmoid.

\subsection{\textcolor{oraclecolor}{OraPO}: Oracle-educated GRPO}\label{sec:OraPO}

\noindent\textbf{Exploration Failure of GRPO in RRG.} 
On radiology report generation, applying vanilla GRPO to a base VLM with limited radiology knowledge often fails to explore effectively. 
The root reason is that the base model carries very limited domain knowledge about radiology. Most base VLMs \cite{DBLP:conf/iclr/ZhengS0WYWL25} are pre-trained on general vision–language corpora and tasks, for example, everyday-image captioning on MS COCO \cite{DBLP:conf/eccv/LinMBHPRDZ14, Wang_2025_ICCV_less_more} and open-domain visual question answering on VQA v2 \cite{DBLP:journals/ijcv/GoyalKASBP19}, rather than medical imaging and reporting. Hence, it tends to miss or misstate rare and clinically challenging findings (\textit{e.g.}, small pneumothorax, subtle interstitial edema, or linear atelectasis) when asked to produce clinical narratives. For many chest X-ray inputs, the sampled group of completions is uniformly low quality with near-zero rewards. Thus, the policy receives little learning signal to drive effective exploration.

Empirically, before any domain-specific fine-tuning, the pretrained backbone attains only $0.034$ macro-F1 (Tab. \ref{tab:ablation}) on clinical effectiveness. As shown in Fig.~\ref{fig:zero_reward_illustration} (left), in the first $50$ optimisation steps approximately $30\%$ of groups have all rewards equal to zero. Under the GRPO construction, such groups give $\bar r \approx 0$ and $\sigma \approx \varepsilon$, so $A_j=\tfrac{r_j-\bar r}{\sigma}\approx 0$ for all $j$.
Consequently, the clipped-ratio term in Eq.~\eqref{eq:grpo} contributes essentially no gradient, and the update is dominated by the KL regularizer to $\pi_{\mathrm{ref}}$ rather than by task reward.
In effect, the optimizer wastes GPU resources by sampling $K$ completions per prompt without extracting useful signals, which both slows convergence and squanders valuable compute.

\noindent\textbf{Zero-Reward Rate (ZRR).}
To resolve GRPO’s exploration failure on RRG, we first \emph{detect} when a prompt yields little or no usable signal, then adapt learning accordingly. We introduce a \emph{Zero-Reward Rate} (ZRR) that quantifies, per prompt, how severely GRPO has stalled.

For prompt $x_i$ with a GRPO group of $K$ completions $\{\hat y_{i,j}\}_{j=1}^{K}$ and rewards $r_{i,j}\in[0,1]$, define the raw zero-reward rate:
\begin{equation}
z_i \;=\; \frac{1}{K}\sum_{j=1}^{K}\mathbf{1}\!\left[r_{i,j}=0\right],
\label{eq:zrr-raw}
\end{equation}
here, $z_i=1$ indicates an all-zero group (no learning signal) and $z_i=0$ indicates at least one non-zero reward (GRPO has exploitable signal).
To reduce step-to-step noise, we maintain an exponential moving average (EMA):
\begin{equation}
\tilde z_i^{(t)} \;=\; \alpha\,\tilde z_i^{(t-1)} \;+\; (1-\alpha)\,z_i^{(t)}, \qquad \alpha\in[0,1),
\label{eq:zrr-ema}
\end{equation}
computed per prompt (or per minibatch) at step $t$.

We then map $\tilde z_i^{(t)}$ to a bounded, monotonic mixing weight that will later control how much oracle education to apply:
\begin{equation}
\begin{aligned}
   w_i^{(t)} =\mathrm{clip}\Big( w_{\min} + (w_{\max}-w_{\min})\big[\tilde z_i^{(t)}\big]^{\gamma},  w_{\min},w_{\max}\Big),
\end{aligned}
\label{eq:zrr-weight}
\end{equation}
where $w_{\min}\!\le\!w_i^{(t)}\!\le\!w_{\max}$ enforces a narrow operating range for stability. The exponent $\gamma$ shapes sensitivity: larger $\gamma$ concentrates education on the most severe failures while leaving ordinary prompts to GRPO.

% Intuitively, if a prompt repeatedly produces all-zero or near-zero groups, $\tilde z_i^{(t)}\!\uparrow$ and $w_i^{(t)}$ increases, signaling stronger need for education; once GRPO begins to surface nonzero rewards, $\tilde z_i^{(t)}\!\downarrow$ and $w_i^{(t)}$ decays, returning emphasis to exploration. 

\noindent\textbf{Integrating Two Objectives.}
With the ZRR-derived weight $w_i^{(t)}$ in hand, we modulate learning per prompt: when GRPO provides usable signal ($w_i^{(t)}$ is small), we emphasise exploration and exploitation; when GRPO stalls ($w_i^{(t)}$ is large), we shift toward direct education via preference alignment.
We combine both signals in a single objective over a batch of size B:
\begin{equation}
\begin{aligned}
\mathcal{L}_{\mathrm{OraPO}}
= \frac{1}{B}\sum_{i=1}^{B}\Big[
\big(1-w_i^{(t)}\big)\,&\mathcal{L}_{\mathrm{GRPO}}(x_i,p_i)
\\ + w_i^{(t)}\,&\mathcal{L}_{\mathrm{DPO}}(x_i,p_i)
\Big],
\end{aligned}
\label{eq:OraPO}
\end{equation}
% \begin{equation}
% \begin{aligned}
% \mathcal{L}_{\mathrm{OraPO}}
% = \mathbb{E}\Big[
% \big(1-w^{(t)}(x,p)\big)\,\mathcal{L}_{\mathrm{GRPO}}(x,p)
% + w^{(t)}(x,p)\,\mathcal{L}_{\mathrm{DPO}}(x,p)
% \Big],
% \end{aligned}
% \label{eq:OraPO}
% \end{equation}
where $\mathcal{L}_{\mathrm{GRPO}}(x_i,p_i)$ is the GRPO loss in Eq.~\eqref{eq:grpo} evaluated on the sampled completions $(x_i,p_i)$, and $\mathcal{L}_{\mathrm{DPO}}(x_i,p_i)$ is the DPO loss in Eq.~\eqref{eq:dpo} computed on preference pairs associated with the same prompt $(x_i,p_i)$. We set the positive to the ground-truth report $y_i^{\star}$ and take negatives from all other model completions:
\begin{equation}
\begin{aligned}
y_i^{+}=y_i^{\star}, \qquad y_i^{-}\in \{\hat y_{i,j}\}_{j=1}^{K} \sim \pi_{\theta}(\cdot \mid x_i,p_i)
\end{aligned}
\label{eq:pairing}
\end{equation}
where $\{\hat y_{i,j}\}_{j=1}^{K}$ is the entire group sampled from $\pi_{\theta}(\cdot\mid x_i,p_i)$.
We then instantiate $\mathcal{L}_{\mathrm{DPO}}(x_i,p_i)$ by applying Eq.~\eqref{eq:dpo} over pairs $\{(x_i,p_i,y_i^{+},y_i^{-})\}$.
This construction reuses low-reward GRPO generation as a preference negative rather than discarding them. As a result, it enables DPO training without requiring any additional hard-negative mining or auxiliary generators. Generated rollouts with zero or low rewards serve as high-quality, domain-specific negative samples that are otherwise difficult to obtain manually.
% The ZRR weight ensures that when the sampled negatives are already close to correct (low zero-reward rate), DPO receives little emphasis, whereas when the group collapses to poor outputs (high zero-reward rate), DPO carries more weight.

Intuitively, \textbf{a high ZRR} flags failure cases where preference \emph{education} must lead. OraPO responds by increasing $w_i^{(t)}$ and applying DPO updates that teach the policy to reject the sampled negatives and align with the ground-truth report. This converts groups into labeled preferences and supplies gradients exactly when GRPO has none/few. \textbf{A low ZRR} indicates that GRPO is uncovering usable reward. This can happen on simpler studies, and it also emerges after DPO has taught the model the cues for difficult findings (\textit{e.g.}, small apical pneumothorax or subtle interstitial edema), enabling a shift from education to exploration and exploitation. In practice, we observe the intended training dynamic, as shown in Fig.~\ref{fig:zero_reward_illustration}: early training is education-heavy with larger ratio of zero-reward groups, higher $w^{(t)}$ is implied, then $w^{(t)}$ gradually declines as the policy learns and GRPO takes over.

\noindent\textbf{What does OraPO Buy Us?}
As shown in Fig.~\ref{fig:zero_reward_illustration}, \textbf{OraPO} drives down the zero-reward ratio faster, and to a lower plateau than na\"ive GRPO, implying fewer wasted rollouts and more gradient-carrying updates. Corroborating this, the class-level performance curves for \emph{Pneumonia} and \emph{Fracture} (clinically challenging, low-prevalence labels) show that \textbf{OraPO} improves \emph{earlier} and sustains \emph{higher} F1 across checkpoints. The coupled trend (rapid ZRR decline \(\rightarrow\) earlier F1 lift on hard classes) indicates that DPO is converting failed exploration into effective supervision rather than merely stabilising optimisation, yielding both better sample efficiency and stronger performance where it matters. \vspace{-1ex}

This initiates a \textbf{self-reinforcing data flywheel}: as the model learns through Oracle education and GRPO’s exploration–exploitation process, it produces increasingly informative negative samples. These, in turn, carry higher reward signals, pushing the model to achieve even greater rewards and completing a positive, compounding feedback cycle.

\subsection{FactScore-based Reward Design}\label{sec:facts_reward}\vspace{-1ex}

OraPO needs a reward that is reliable and efficient to compute. Many RL-based RRG approaches optimize report-level proxies or embedding similarity, which favors fluent paraphrases over verifiable clinical facts and can yield plausible but clinically incorrect narratives. We introduce a FactScore-based reward (\textbf{FactS}) that treats the report as its own rationale: extract atomic clinical facts and test each against the ground-truth label set to produce a dense, fact-entailment signal for reinforcement learning.

\noindent\textbf{Step 1: Extract atomic clinical facts.}
Inspired by FactScore \cite{DBLP:conf/emnlp/MinKLLYKIZH23}, we define a \emph{fact} as an atomic, verifiable clinical statement (e.g.\ finding, attribute, and optionally location), for instance,  ``no pleural effusion'', ``linear atelectasis at the left base''.
Given a generated report $\hat y_i$, we prompt GPT-4.1~\cite{DBLP:journals/corr/abs-2303-08774} to extract a set of atomic facts:
\begin{equation}
\mathcal{F}(\hat y_i)=\{s_{i,k}\}_{k=1}^{K_i},
\label{eq:extact_atomic_facts}
\end{equation}
ensuring each $s_{i,k}$ is concise and verifiable.

\noindent\textbf{Step 2: Per-label entailment against ground-truth.}
Let $\mathcal{L}$ be the fixed label set (\textit{e.g.}\ 14 CheXpert labels) and let $z^{\ast}_{i,\ell}\in\{0,1\}$ be the ground-truth for label $\ell\in\mathcal{L}$ on study $x_i$.
We then run an LLM-based entailment check that maps the fact set to label-level predictions:
\begin{equation}
\hat z_{i,\ell} \;=\; \mathbf{1}\!\left[\exists\, s\in\mathcal{F}(\hat y_i)\ \text{s.t.}\ \texttt{entails}(s,\ell)=\texttt{true}\right]
\end{equation}\label{eq:fact_to_label_pred}
for all $\ell\in\mathcal{L}$. Explicit contradictions to $\ell$ are treated as false positives for that label (discouraging unsupported claims).

\noindent\textbf{Step 3: A dense, interpretable reward.}
Based on $(\hat z_{i,\ell},z^{\ast}_{i,\ell})_{\ell\in\mathcal{L}}$, we compute standard per-instance precision $P_i$ and recall $R_i$, and the reward as an $F_{\beta}$ score
\begin{equation}
r(x_i,\hat y_i)
=
F_{\beta,i}
=
\frac{(1+\beta^{2})\,P_i R_i}{\beta^{2} P_i + R_i + \xi},
\label{eq:facts-reward}
\end{equation}
with $\beta>1$ placing more weight on recall (penalizing missed positive findings) and $\beta<1$ emphasizing precision; $\xi$ is a small constant for stability.
The reward is dense and label-wise interpretable, directly supervising diagnostic correctness while remaining token-efficient. It requires no critic, no auxiliary reward model, nor long reasoning traces.
By anchoring each sentence to specific clinical facts and aligning them with ground-truth labels, the FactS Reward reduces unsupported claims and improves coverage of required findings, providing a faithful signal for OraPO.

\begin{table}[t]
\centering
\caption{Experimental results on the CheXpert Plus dataset \cite{DBLP:journals/corr/abs-2405-19538} using \textbf{task-specific RRG algorithms}. We report macro-averaged \textbf{Precision}, \textbf{Recall}, and \textbf{F1} across 14 CheXpert pathologies. \textbf{Train Size} is the number of training samples. Some baselines train on multiple corpora and/or in multiple stages. Best and second best are in \textbf{bold} and \underline{underline}, respectively. \vspace{-2ex}}
\small
\resizebox{\linewidth}{!}{ 
\begin{tabular}{l|l|c|l}
\toprule
\textbf{Algorithm} & \textbf{Venue} & \textbf{Precision, Recall, F1}  & \textbf{Train Size} \\
\midrule
R2GenRL \cite{DBLP:conf/acl/Qin022} & ACL22      & 0.193,\;0.229,\;0.196  & 223K \\
XProNet \cite{DBLP:conf/eccv/WangBH22} & ECCV22     & 0.314,\;0.247,\;0.259  & 223K \\
MSAT \cite{DBLP:conf/miccai/WangTWLZ22}       & MICCAI22   & 0.044,\;0.142,\;0.057  & 223K \\
ORGan \cite{DBLP:conf/acl/HouXCLL23}      & ACL23      & 0.288,\;0.287,\;0.277  & 223K \\
M2KT \cite{DBLP:journals/mia/YangWGZZX23}       & MIA21      & 0.044,\;0.142,\;0.058  & 223K \\
TIMER \cite{DBLP:conf/chil/WuH023}       & CHIL23     & \underline{0.345},\;0.238,\;0.234 & 223K \\
CvT2DistilGPT2 \cite{DBLP:journals/artmed/NicolsonDK23} & AIM23 & 0.285,\;0.252,\;0.246 & 223K \\
R2Gen \cite{DBLP:conf/emnlp/ChenSCW20}     & EMNLP20    & 0.318,\;0.200,\;0.181  & 223K \\
R2GenCMN \cite{DBLP:conf/acl/ChenSSW20} & ACL21    & 0.329,\;0.241,\;0.231  & 223K \\
Zhu et al. \cite{DBLP:conf/miccai/ZhuMMPSL23} & MICCAI23 & 0.217,\;{0.308},\;0.205  & 223K \\
CAMANet \cite{DBLP:journals/titb/WangBYSH24} & IEEE JBHI23 & 0.328,\;0.224,\;0.216 & 223K \\
ASGMD \cite{DBLP:journals/eswa/XueTTQX24}      & ESWA24     & 0.146,\;0.108,\;0.055  & 223K \\
Token-Mixer \cite{DBLP:journals/tmi/YangYFZYWJLHH24} & IEEE TMI23 & 0.309,\;0.270,\;{0.288}  & 223K \\
PromptMRG \cite{DBLP:conf/aaai/JinCLC24} & AAAI24   & 0.258,\;0.265,\;0.281  & 223K \\
R2GenGPT \cite{WANG2023100033} & Meta-Rad.23 & 0.315,\;0.244,\;0.260  & 223K \\
WCL \cite{DBLP:conf/emnlp/YanHLDCGMH21}          & EMNLP21    & 0.335,\;0.259,\;0.256  & 223K \\
R2GenCSR \cite{DBLP:journals/corr/abs-2408-09743} & arXiv24  & 0.315,\;0.247,\;0.259 & 223K \\
Wang et al. \cite{DBLP:journals/visintelligence/WangLWJRJLT25} & VIS INT25 & 0.175,\;0.099,\;0.078 & 223K \\
CheXagent \cite{DBLP:journals/corr/abs-2401-12208} & arXiv24 & 0.373,\;0.194,\;0.186 & 8.5M \\ 
MambaXray-B \cite{DBLP:conf/cvpr/WangWLMW0025} & CVPR25         & 0.333,\;0.264,\;0.273  & 1.27M \\
MambaXray-L \cite{DBLP:conf/cvpr/WangWLMW0025} & CVPR25        & \textbf{0.377},\;\underline{0.319},\;\underline{0.335}  & 1.27M \\
VLCI \cite{DBLP:journals/tip/ChenLWZLLL25}       & IEEE TIP25    & 0.341,\;0.175,\;0.163 & 223K \\
\midrule 
\textbf{Ours } & --        & 0.237,\;\textbf{0.832},\;\textbf{0.341}  & \textbf{1K} \\
\bottomrule
\end{tabular}
}
\label{tab:rrg_task_specific_compare}
\end{table}

\begin{table}[t]
\centering
\caption{Experimental results on the CheXpert Plus benchmark \cite{DBLP:journals/corr/abs-2405-19538} across diverse \textbf{LLMs/VLMs}, all supervised fine-tuned with R2GenGPT~\cite{WANG2023100033} on CheXpert Plus. We report macro-averaged \textbf{Precision}, \textbf{Recall}, and \textbf{F1} across 14 CheXpert pathologies. The \textbf{Params} listed denotes the parameters that need to be tuned in the training phase. \textbf{FT-Size} indicates the number of samples used for fine-tuning. Best and second best are in \textbf{bold} and \underline{underline}, respectively. \vspace{-2ex}}
\small
\resizebox{\linewidth}{!}{ 
\begin{tabular}{l|l|c|l|l}
\toprule
\textbf{LLM/VLM} & \textbf{Year} & \textbf{Precision, Recall, F1} & \textbf{Params} & \textbf{FT-Size} \\
\midrule
Vicuna-V1.5~\cite{DBLP:conf/nips/ZhengC00WZL0LXZ23} & 2023 & 0.334,\;0.258,\;{0.276} & 6.7B & 223K\\
Qwen-1.5~\cite{DBLP:journals/corr/abs-2309-16609}        & 2024 & 0.303,\;0.233,\;0.241 & 7.7B & 223K\\
Qwen-2~\cite{DBLP:journals/corr/abs-2309-16609}          & 2024 & 0.313,\;{0.269},\;0.261 & 7.6B & 223K\\
InternLM~\cite{DBLP:journals/corr/abs-2403-17297}   & 2024 & 0.307,\;\underline{0.274},\;\underline{0.284} & 7.3B & 223K\\
Llama-2 (7B)~\cite{DBLP:journals/corr/abs-2307-09288}  & 2023 & 0.315,\;0.244,\;0.260 & 6.7B & 223K\\
Llama-2 (13B)~\cite{DBLP:journals/corr/abs-2307-09288} & 2023 & 0.321,\;0.254,\;0.267 & 13.0B & 223K\\
Llama-3~\cite{DBLP:journals/corr/abs-2407-21783}     & 2024 & 0.306,\;0.232,\;0.222 & 8.0B & 223K\\
Llama-3.1~\cite{DBLP:journals/corr/abs-2407-21783}   & 2024 & 0.295,\;0.251,\;0.242 & 8.0B & 223K\\
GPT2-Medium~\cite{radford2019language}  & 2019 & \textbf{0.358},\;0.186,\;0.165 & 354M & 223K\\
Orca-2 (7B)~\cite{DBLP:journals/corr/abs-2311-11045}   & 2023 & 0.330,\;0.251,\;0.271 & 6.7B & 223K\\
Orca-2 (13B)~\cite{DBLP:journals/corr/abs-2311-11045}  & 2023 & 0.317,\;0.242,\;0.257 & 13.0B & 223K\\
Deepseek-LLM~\cite{DBLP:journals/corr/abs-2401-02954} & 2024 & \underline{0.336},\;0.256,\;0.253 & 6.9B & 223K\\
Yi-1.5 (6B)~\cite{DBLP:journals/corr/abs-2403-04652}        & 2024 & 0.322,\;0.229,\;0.226 & 6.1B & 223K\\
Yi-1.5 (9B)~\cite{DBLP:journals/corr/abs-2403-04652}        & 2024 & \underline{0.336},\;0.241,\;0.243 & 8.8B & 223K\\
InternVL-2~\cite{DBLP:journals/corr/abs-2403-17297} & 2023 & 0.196,\;0.127,\;0.132 & 8.0B & 223K\\
MiniCPM-V2.5~\cite{DBLP:journals/corr/abs-2408-01800} & 2024 & 0.254,\;0.152,\;0.122 & 8.4B& 223K\\
\midrule 
\textbf{Ours} & -- & 0.237,\;\textbf{0.832},\;\textbf{0.341} & \textbf{3B} & \textbf{1K}\\
\bottomrule
\end{tabular}
}
\label{tab:LLMs_compare}
\end{table}

\begin{table}[t]
\centering
\caption{
Experimental results on the MIMIC-CXR dataset  \cite{DBLP:journals/corr/abs-2405-19538}. We report macro-averaged \textbf{Precision}, \textbf{Recall}, and \textbf{F1} across 14 CheXpert pathologies. \textbf{Train Size} is the number of training samples. Some baselines train on multiple corpora and/or in multiple stages. Best are in \textbf{bold}. \vspace{-2ex}}
\small
\resizebox{\linewidth}{!}{
\begin{tabular}{l|l|c|l}
\toprule
\textbf{Algorithm} & \textbf{Venue} & \textbf{Precision, Recall, F1} & \textbf{Train Size} \\
\midrule
R2Gen \cite{DBLP:conf/emnlp/ChenSCW20} & EMNLP20 & 0.333, 0.273, 0.276 & 227K \\
CXR-RePaiR \cite{DBLP:conf/ml4h/EndoKKNR21} & ML4H21 & \ \ \ -- \ \ \ , \ \ \ -- \ \ \ , 0.274 & 227K  \\
MRG/SILC \cite{DBLP:journals/tmi/LiuGYX24} & TIP24 & \textbf{0.457}, 0.337, 0.330 & 227K  \\
CXRMate \cite{nicolson2024longitudinaldata} & IMU24 & 0.277, 0.351, 0.283 & 227K \\
HERGen \cite{DBLP:conf/eccv/WangDY24} & ECCV24 & 0.415, 0.301, 0.317 & 145K \\
ChEX \cite{DBLP:conf/eccv/MullerKR24} & ECCV24 & \ \ \ -- \ \ \ , \ \ \ -- \ \ \ , 0.326 & 227K \\
MambaXray-L \cite{DBLP:conf/cvpr/WangWLMW0025} & CVPR25 & 0.371, 0.321, 0.340 &  1.27M \\
GIT-CXR \cite{sirbu2025git} & INFO25 & \ \ \ -- \ \ \ , \ \ \ -- \ \ \ , 0.327 & 227K \\
MCA-RG \cite{DBLP:conf/miccai/XingSZFYY25} & MICCAI25 & 0.443, 0.306, 0.335 & 227K \\
\midrule 
\textbf{Ours} & -- & 0.242, \textbf{0.891}, \textbf{0.357} & \textbf{1K} \\
\bottomrule
\end{tabular}
}
\label{tab:mimic_macro}
\end{table}

\begin{table}[t]
\centering
\caption{Ablation on CheXpert Plus \cite{DBLP:journals/corr/abs-2405-19538}: Impact of the \textbf{FactS} reward and \textbf{OraPO}. The first row indicates the base Qwen2.5-VL-3B-Instruct \cite{DBLP:journals/corr/abs-2502-13923} direct test results, without any fine-tuning. The second row corresponds to a naïve GRPO-based RLHF baseline using accuracy as the reward \cite{guo2025deepseeknature}. We report per-label \textbf{Precision}, \textbf{Recall}, and \textbf{F1} across 14 findings. Best are in \textbf{bold}.\vspace{-2ex}}
\small
\resizebox{\linewidth}{!}{
\begin{tabular}{cccc|c|l}
\toprule
\textbf{FactS} & \textbf{GRPO} & \textbf{DPO}  & \textbf{SFT} & \textbf{Train Size} & \textbf{Precision, Recall, F1} \\
\midrule
 $\times$ & $\times$ & $\times$ & $\times$ & 0  & 0.097,\;\;0.104,\;\;0.034 \\
 $\times$ & $\checkmark$ & $\times$ &  $\times$ & 1,000  & 0.026,\;\;0.162,\;\;0.089 \\
$\checkmark$ & $\checkmark$ & $\times$ &  $\times$ & 1,000 & 0.204,\;\;0.605,\;\;0.291 \\
 $\checkmark$ & $\checkmark$ & $\checkmark$ &  $\times$ &  400  & {0.217},\;\;0.732,\;\;0.296 \\
  $\checkmark$ & $\checkmark$ &  $\times$ & $\checkmark$ &  1,000  & {0.171},\;\;0.176,\;\;0.106  \\
 \midrule
 $\checkmark$ & $\checkmark$&  $\checkmark$ &  $\times$ &  1,000  & \textbf{0.237},\;\;\textbf{0.832},\;\;\textbf{0.341} \\
\bottomrule
\end{tabular}
}
\label{tab:ablation}
\end{table}

\begin{table}[t]
\centering
\caption{Experimental results on CheXpert validation set with gold labels from certified radiologists \cite{DBLP:conf/aaai/IrvinRKYCCMHBSS19}. \vspace{-1ex}}
\small
\resizebox{0.9\linewidth}{!}{
\begin{tabular}{l|c|l}
\toprule
\textbf{Method}  & \textbf{Train Size} & \textbf{Precision, Recall, F1} \\
\midrule
GPT4.1~\cite{DBLP:journals/corr/abs-2303-08774}                        & --  & {0.292}, 0.305, 0.253 \\
GPT5 (Thinking) \cite{openai2025gpt5systemcard}                        & --  & 0.261, 0.421, 0.301 \\
\midrule
CheXagent \cite{DBLP:journals/corr/abs-2401-12208}                       & 8.5M  & {0.462}, 0.226, 0.254 \\
MambaXray-B \cite{DBLP:conf/cvpr/WangWLMW0025}                  & 1.27M & 0.362, 0.254, 0.238 \\
MambaXray-L \cite{DBLP:conf/cvpr/WangWLMW0025}                 & 1.27M & 0.395, 0.326, 0.280 \\
\midrule
Ours                            &   1K    &  0.234, {0.641}, {0.288} \\
% Ours (CKPT1350)            &     GT Rerports + CheXpert Labels                    &   1K    &  0.190, 0.909, 0.279 \\
% Ours             &       GT Rerports + GPT41 Labels                  &   1K    & 0.197, 0.579, 0.276 \\
\bottomrule
\end{tabular}
}
\label{tab:gold_label_compare}
\end{table}

\section{Experiments}
\subsection{Experimental Setup}\vspace{-1ex}

\noindent\textbf{Datasets.} CheXpert Plus \cite{DBLP:journals/corr/abs-2405-19538} is a recent large-scale paired image–text resource that links chest X-ray DICOMs to fully de-identified radiology reports, organized into report subsections and annotated for 14 target pathologies (i.e.\ disease labels). 
The public release includes $\sim$223K image–report pairs across $\sim$187.7K studies and $\sim$64.7k patients, totaling $\sim$36M report tokens. MIMIC-CXR \cite{DBLP:journals/corr/abs-1901-07042} is a large, publicly available paired image–text resource of chest X-rays in DICOM format with contemporaneous, de-identified radiology reports. It contains 377,110 images from 227,943 imaging studies covering 65,379 patients. Each study includes one or more views (typically frontal and/or lateral) and a free-text report that often follows Findings and Impression sections. For data-efficient fine-tuning, we sample 1{,}000 studies per dataset (\textit{i.e.} 1{,}000 from CheXpert-Plus and 1{,}000 from MIMIC-CXR) with a balanced ground-truth label set via CheXpert labeller \cite{DBLP:conf/aaai/IrvinRKYCCMHBSS19}. We also exclude reports with missing sections. Following recent works \cite{DBLP:conf/cvpr/WangWLMW0025}, we report results on the test split and extract labels from both generated and GT reports by the CheXpert labeller. We further evaluate on the CheXpert Plus human gold set, whose expert-verified labels provide a more reliable basis for evaluation and stronger conclusions.

\noindent\textbf{Baselines.} Following the recent benchmarking study \cite{DBLP:conf/cvpr/WangWLMW0025}, we compare our proposed approach against a broad suite of methods: \textbf{28 task-specific RRG models} spanning diverse architectures and \textbf{16 open-source LLMs/VLMs}, as listed in Tables \ref{tab:rrg_task_specific_compare}, \ref{tab:LLMs_compare} and \ref{tab:mimic_macro}.
All task-specific algorithms are fine-tuned on at least the full training set of the dataset used in each experiment (CheXpert Plus or MIMIC-CXR), with some methods additionally incorporating multi-source corpora. Thus, our 1K training set sampled from CheXpert Plus, represents \(\sim\)0.45\% of the standard 223K split; and only \(\sim\)0.079\% of the 1.27M used by MambaXray \cite{DBLP:journals/tip/ChenLWZLLL25}, which is the current SOTA model. A similar ratio applies to MIMIC-CXR.

\noindent\textbf{Evaluation Metric.} We follow \cite{DBLP:conf/cvpr/WangWLMW0025} to report \emph{clinical effectiveness} as macro-averaged over labels: for each of the 14 conditions we compute Precision, Recall, and F1 on the test set, then take the unweighted mean across labels so frequent and rare conditions contribute equally. \vspace{-2ex}

% \noindent\textbf{Implementation Details.}
% We fine-tuned a 3-billion-parameter Qwen-2.5-VL model using a small subset of CheXpert Plus (144 image–report pairs) without additional supervised pre-training or alignment stages, following a reinforcement-learning-with-verifiable-rewards setup. The implementation uses Hugging Face TRL’s GRPOTrainer with transformers and accelerate. Experiments ran on four NVIDIA A10 GPUs (24 GB each) under PyTorch 2.7.1 and CUDA 12.1. To keep GRPO as the primary learning signal while still correcting hard, zero-reward cases, we bound the DPO educator's mixing weight to a narrow interval, by setting $w_{\min}=0.05,\; w_{\max}=0.15,\; \gamma=2.$ Thus \(w_t \to w_{\max}\) only as the zero-\(F_\beta\) rate approaches \(1\) (no usable GRPO signal), and \(w_t\) quickly decays toward \(w_{\min}\) once GRPO yields non-zero rewards, ensuring GRPO dominates ordinary updates while DPO remains a light, targeted aid. Regarding the objective function implementation, we adopt its improved vairants of Dr-GRPO  and a variant of DPO,  removing the length bias of different reports. 

\paragraph{Implementation Details.}
We fine-tune a \textbf{3B} Qwen-2.5-VL policy \cite{DBLP:journals/corr/abs-2502-13923} on the \textbf{1{,}000} CheXpert Plus studies in an \textbf{RL-only} setting (no SFT/alignment), using Hugging Face \texttt{TRL}’s \texttt{GRPOTrainer} with  \texttt{transformers} and  \texttt{accelerate}. Training runs on four NVIDIA A10 GPUs (24\,GB) with PyTorch~2.7.1/CUDA~12.1. We empirically set batch size $B=16$. For each prompt, we sample $K=8$ completions, compute the \textbf{FactS} reward per completion, and optimize the objective in Eq.~\eqref{eq:OraPO}. More detailed description of the hyper-parameters for OraPO can be found in the Supplementary.

% GRPO provides the default learning signal, while DPO is invoked adaptively via the ZRR schedule. To prioritize GRPO yet correct hard, zero-reward cases, we map the EMA zero-reward rate $\tilde z_i^{(t)}$ with $\alpha=0.5$ in Eq. \eqref{eq:zrr-ema} to a bounded mixing weight $w_i^{(t)}$ in Eq.~\eqref{eq:zrr-weight} with $w_{\min}\!=\!0.05,\quad w_{\max}\!=\!0.15,\quad \gamma\!=\!2.$ Thus, $w_i^{(t)}\!\to\!w_{\max}$ only when a prompt’s ZRR approaches 1 (no usable GRPO signal), and it {quickly decays} toward $w_{\min}$ once GRPO yields non-zero rewards, ensuring GRPO dominates ordinary updates and DPO remains a light, targeted aid. We adopt two length-aware variants to handle variable-length reports: DR.GRPO \cite{DBLP:journals/corr/abs-2503-20783} to mitigate length bias in on-policy updates, and {LN-DPO}~\cite{DBLP:conf/naacl/AhrabianLPCBPS25} to normalize preference margins by sequence length. 

\subsection{Main Results and Analysis on CheXpert Plus}

\paragraph{Against RRG task-specific algorithms.} %We evaluate OraPO on CheXpert Plus using only 1K training studies and compare with specialized RRG approaches that typically fine-tune on the full 223K split or even multi-source corpora. 
As shown in Table~\ref{tab:rrg_task_specific_compare}, \textbf{OraPO+FactS} sets the new SOTA \textbf{F1 = 0.341 (best)} and \textbf{Recall = 0.832 (best)} on CheXpert Plus, significantly surpassing best-performing baselines by \(\mathbf{+160.8\%}\) in recall. When compared with the most recent baseline, VLCI, our proposed method outperforms VLCI by \(\mathbf{+109.2\%}\) in F1 and \(\mathbf{+375.4\%}\) in recall. While our precision trails MambaXray-L, this is a deliberate trade-off between emphasizing per-finding coverage and accurate statements of detected findings. In practice, recall-oriented drafts are preferred. Since radiologists perform the final review, they tend to favour AI assistance that has higher sensitivity (even at the cost of precision) \cite{hendrix2022radiologist}. Overall, our method sets the new SOTA performance in F1 and recall, with \textbf{three orders of magnitude} less training data.

\noindent \textbf{Against Alignment-based SFT for LLM/VLM.} In this experiment, we compare our framework against R2GenGPT~\cite{WANG2023100033}, a popular alignment-based SFT approach that is widely used for training large LLM/VLM. We report the performance of R2GenGPT~\cite{WANG2023100033} with different LLM/VLM backbones in Table~\ref{tab:LLMs_compare}. Our approach is trained with a compact \textbf{3B} policy on \textbf{1K} CheXpert Plus samples. Under this small-data setup, \textbf{OraPO} attains \(\mathbf{F1 = 0.341}\) and \(\mathbf{Recall = 0.832}\), significantly outperforming larger aligned backbones trained on 223K.
% : F1 improves over InternLM by \(\mathbf{+20.1\%}\) and over Qwen-2 (0.261) by \(\mathbf{+30.7\%}\); recall improves over InternLM by \(\mathbf{+203.6\%}\) and over GPT-2 Medium by \(\mathbf{+347.3\%}\).
These results validate that our proposed \textbf{OraPO} with fact-level supervision is a superior learning paradigm to standard vision–language alignment, achieving higher performance with \textbf{less than half backbone parameters} and \textbf{markedly less training data}.

\noindent \textbf{Evaluation on Human Gold Label Set}. To obtain a more faithful evaluation, we assess models on the CheXpert validation set with human gold labels: 200 studies from 200 patients independently annotated by three board-certified radiologists, where a majority vote defines the ground truth  \cite{DBLP:conf/aaai/IrvinRKYCCMHBSS19}. Table \ref{tab:gold_label_compare} reports macro Precision/Recall/F1 across the 14 findings. With only 1K training cases, our method reaches 0.234/0.641/0.288, surpassing MambaXray-B,  MambaXray-L and CheXagent in both Recall and F1. These gains arrive while using 0.1\% of MambaXray’s data and around 0.01\% of CheXagent’s.
% The human-verified labels remove the dependency on the CheXpert labeller used by many works to extract “ground-truth” from reports, thereby \textbf{avoiding extraction biases and better reflecting clinical correctness}. Under this stricter setting, our approach maintains state-of-the-art recall and competitive F1, indicating substantially fewer missed findings.

We also evaluate two commercial APIs (GPT-4.1~\cite{DBLP:journals/corr/abs-2303-08774} and GPT-5 Thinking \cite{openai2025gpt5systemcard}) on the CheXpert validation set with human gold labels, using the identical image input and report-generation prompt. GPT-5 is configured with a medium reasoning level. As reported in Table~\ref{tab:gold_label_compare}, our model outperforms GPT-4.1 by +110.2\% in recall and +13.8\% in F1. Relative to GPT-5 Thinking, our model achieves +52.3\% higher recall, though F1 is -4.3\% lower. Critically, compared to GPT-5 Thinking, our compact 3B model delivers real-time inference (3.3 s/image vs. 25.2 s/image), a lightweight and transparent footprint (3B vs. trillions of parameters), and cheaper deployment that eliminates per-token API fees. These advantages make our method substantially more cost-efficient and suitable for time-critical clinical use than calling commercial APIs. Overall, the results confirm that our improvements persist when measured against expert gold labels, while being dramatically more data-efficient.

\subsection{Main Results and Analysis on MIMIC-CXRs}
On MIMIC-CXR, our approach attains F1 = 0.357 and Recall = 0.891 using just 1K training samples, as shown in Table \ref{tab:mimic_macro}. This yields a +5.0\% F1 gain over the best baseline (0.340, MambaXray-L) and a +153.8\% recall gain over the strongest baseline recall (0.351, CXRMate). The markedly higher recall under macro averaging indicates substantially fewer omissions across both common and rare pathologies, aligning with clinical preference for high sensitivity and omission avoidance \cite{DBLP:conf/naacl/MiuraZTLJ21, seyyed2021underdiagnosis, rao2024rexerr, DBLP:journals/corr/abs-2511-13361,  noguchi2023analysis}. At the same time, the overall F1 is the best among all baselines, despite a precision of 0.242. Together with the CheXpert-Plus results, these findings show that OraPO generalises across datasets and achieves state-of-the-art clinical effectiveness with 2--3 orders-of-magnitude less data than prior approaches.

\subsection{Ablation Study} To study the effectiveness and efficiency of our proposed method, we evaluate four model variants on CheXpert Plus using macro \textbf{Precision}, \textbf{Recall}, and \textbf{F1}. As reported in Table~\ref{tab:ablation}, the second row is a na\"ive GRPO-based RLHF baseline that uses \emph{accuracy} as reward (as in DeepSeek-R1 \cite{guo2025deepseeknature}), which yields only marginal gains over the base model without any domain-specific training (row 1). Adding \textbf{FactS} to GRPO (row 3) lifts F1 from 0.089 to 0.291 \((+227\%)\), Recall from 0.162 to 0.605 (+274\%), and Precision from 0.026 to 0.204 (+685\%), showing that fact-level entailment provides dense, clinically aligned supervision. Layering \textbf{OraPO} on top (row 6) further improves F1 to 0.341 (+17.2\% vs.\ \textbf{FactS}-only), with Recall improved by (+37.5\%) and Precision improved by (+16.2\%). Notably, \textbf{OraPO+FactS} with only \textbf{400} studies (row 4) attains F1 \(=\) 0.296, Precision $=$ 0.217 and Recall \(=\) 0.732—--surpassing the \textbf{FactS}-only 1K setting (F1 \(=\) 0.291, Precision $=$ 0.204, and Recall \(=\) 0.605)--—providing evidence of OraPO's superb \emph{data efficiency}.  We further observe that the \textbf{GRPO+SFT} variant (with FactS) collapses to very low recall (0.176) and F1 (0.106), a \textbf{-70.9\%} and \textbf{-63.6\%} drop vs.\ FactS+GRPO, respectively. The core issue here is that plain SFT only imitates ground-truth prose: the model learns “what to say when it’s right,” but not “what not to say,” so reports that omit all true positives aren’t explicitly discouraged, leading to recall and F1 collapses. Beyond that, SFT on small datasets is known to be unstable and sample-inefficient: fine-tuning with limited data often over-fits, degrading generalisation \cite{DBLP:conf/emnlp/PecherCBSSB24, DBLP:conf/acl/Du023}. OraPO fixes this by explicitly pushing probability mass away from ``no-TP'' outputs such that the recall is effectively boosted when trained with limited set. Overall, \textbf{FactS} supplies faithful, sentence-level rewards, while \textbf{OraPO} converts zero-reward groups into preference updates, yielding higher precision and recall and faster convergence on scarce yet clinically important signals.

% Measuring the Instability of Fine-Tuning

\section{Conclusion}
In this work, we present \textbf{OraPO}, an oracle-educated RL training algorithm for high-quality radiology report generation under tight data and compute budgets. To our knowledge, we are the \emph{first} to \emph{couple DPO with GRPO}: when GRPO exploration fails, we recycle those rollouts as negatives and apply a lightweight DPO step to recover useful gradients. In parallel, we design the \textbf{FactS} reward to provide dense, fact-level, clinically faithful supervision tailored to RRG. Comprehensive experiments show that \textbf{OraPO sets the new SOTA performance while using 2–3 orders of magnitude fewer training samples}. In future work, we target four directions: (1) Extend OraPO beyond RRG to sparse-signal tasks (\textit{e.g.} multi-step math, code) where vanilla GRPO struggles. (2) Study scaling across model size and data/training budgets beyond our current 3B/1K setup. (3) Tackle non-verifiable rewards via rubric-driven signals (\textit{e.g.} Rubrics as Rewards \cite{gunjal2025rubricsrewardsreinforcementlearning}). (4) Enrich FactS with fine-grained attribute entailment (location, laterality, size/severity, temporal change) to reduce diagnostic attribute misalignment.

\clearpage
\appendix

\twocolumn[{%
\centering
% {\Large \bfseries \par}
{\Large \bfseries Supplementary Material of \textcolor{oraclecolor}{OraPO}: Oracle-educated Reinforcement Learning \par}
{\Large \bfseries for Data-efficient and Factual Radiology Report Generation\par}
\vspace{1.0em}
}]

% %%%%%%%%% AUTHORS - PLEASE UPDATE
% \author{Zhuoxiao Chen$^{1,2\dagger}$, Hongyang Yu$^{1\star}$, Ying Xu$^{1}$, Yadan Luo$^{2}$, Long Duong$^1$, Yuan-Fang Li$^{1\star}$ \\
% $^1$Oracle Health \& AI, 
% $^2$The University of Queensland\\
% \tt\small\{ivan.chen, hongyang.yu, ying.x.xu, long.duong, yuanfang.li\}@oracle.com \\ \tt\small\{zhuoxiao.chen, y.luo\}@uq.edu.au
% \vspace{-1ex}
% }

% \begin{document}
% \maketitle
% % \let\thefootnote\relax\footnotetext{$^\dagger$ Intern of Oracle, $^{\star}$ Corresponding author.}

% \twocolumn[{%
% \renewcommand\twocolumn[1][]{#1}
% \maketitle
% }]

% % --- author footnote under the title, full width ---
% \begingroup
% \renewcommand\thefootnote{}% no mark
% \footnotetext{$^\dagger$ Work done during an internship at Oracle Health \& AI. $^{\star}$ Joint senior authors.}
% \addtocounter{footnote}{-1}% keep numbering correct
% \endgroup
% ---------------------------------------------------

\noindent This supplementary material includes:
\begin{itemize}
\item \textbf{Algorithm \ref{alg:orapo}:} A detailed, step-by-step description of the proposed method.
\item \textbf{Section \ref{sec.implement_details}:} Comprehensive implementation details.
\item \textbf{Section \ref{sec.mimiccxr_micro}:} Additional experimental results and analysis on the MIMIC-CXR dataset.
\item \textbf{Section \ref{sec.qualitative}:} Qualitative studies, including examples of generated reports and assessments on extracted facts.
\end{itemize}

\section{More Implementation Details} \label{sec.implement_details}
Table~\ref{tab:zrr_hyper} summaries the hyperparameters of our method and the ranges explored during tuning. We use a small effective batch size of $B=16$, chosen to fit our compute-efficient 4$\times$A10 setup, and a conservative learning rate of $2.5\times10^{-7}$ to stabilise GRPO updates under high reward variance. The GRPO group size is set to $K=8$, which we found to provide a good balance between exploration of diverse rollouts and stable gradient estimates. For the ZRR-controlled mixing, we tune the EMA momentum $\alpha$ in ${0.4, 0.5, 0.6}$ and select $\alpha=0.5$ so that the zero-reward rate reacts quickly enough to persistent failures, while still smoothing out step-to-step noise. The DPO mixing weights are constrained to a relatively narrow band, with $w_{\min}=0.05$ and $w_{\max}=0.15$. In practice, we observed that larger $w_{\max}$ values cause DPO to dominate the update, which slows GRPO exploration and makes the overall behavior resemble supervised fine-tuning rather than reinforcement learning. Conversely, keeping a non-zero $w_{\min}$ ensures that DPO remains available as a gentle corrective signal even when GRPO rewards are mostly informative. Finally, we choose a relatively sharp exponent $\gamma=2.0$ for mapping ZRR to the mixing weight. This choice makes the weight rise toward $w_{\max}$ only when a prompt experiences many zero-reward groups, so DPO takes over in genuinely hard, unlearned cases, but the weight quickly falls back toward $w_{\min}$ once non-zero rewards appear, allowing GRPO to dominate ordinary updates. We adopt two length-aware variants to handle variable-length reports: DR.GRPO \cite{DBLP:journals/corr/abs-2503-20783} to mitigate length bias in on-policy updates, and {LN-DPO}~\cite{DBLP:conf/naacl/AhrabianLPCBPS25} to normalise preference margins by sequence length.

\begin{table}[t]
\centering
\caption{Hyperparameter settings of the proposed method (selected in \underline{underline}).}
\small
\resizebox{\linewidth}{!}{
\begin{tabular}{l|l|l}
\toprule
\textbf{Parameter} & \textbf{Description} & \textbf{Search Range / Setting} \\
\midrule
$B$ & Effective Batch Size &
$\{\underline{16}\}$ \\
LR & Learning Rate &
$\{1e-6, \underline{2.5e-7}\}$ \\ 
$K$ & GRPO sampled group size & $\{4, \underline{8}, 16\}$ \\
$\alpha$ & EMA momentum for zero-reward rate (ZRR) &
$\{0.4, \underline{0.5}, 0.6\}$ \\
$w_{\min}$ & Minimum DPO mixing weight & $\{0.02, \underline{0.05}, 0.1\}$\\
$w_{\max}$ & Maximum DPO mixing weight &
$\{\underline{0.15}, 0.3\}$ \\
$\gamma$ & Sharpening exponent for mapping ZRR to $w_i^{(t)}$ &
$\{1.0, \underline{2.0}\}$ \\
\bottomrule
\end{tabular}
}
\label{tab:zrr_hyper}
\end{table}

% ========= Micro =========
\begin{table}[t]
\centering
\caption{Experimental results (micro averaging) on the MIMIC-CXR dataset \cite{DBLP:journals/corr/abs-2405-19538}.}
\small
\resizebox{\linewidth}{!}{
\begin{tabular}{l|l|c|l}
\toprule
\textbf{Algorithm} & \textbf{Venue} & \textbf{Precision, Recall, F1} & \textbf{Train Size} \\
\midrule
MedRAT \cite{DBLP:conf/eccv/HirschDT24} & ECCV24 & 0.285, 0.265, 0.227 & 223K \\
MET~\cite{DBLP:conf/cvpr/WangLWZ23} & CVPR23 & 0.364, 0.309, 0.311 & 223K \\
KiUT~\cite{DBLP:conf/cvpr/Huang0Z23} & CVPR23 & 0.371, 0.318, 0.321 & 369K \\
DCL~\cite{DBLP:conf/cvpr/0006LCLLC23} & CVPR23 & 0.471, 0.352, 0.373 & 223K\\
RGRG~\cite{DBLP:conf/cvpr/TanidaMKR23} & CVPR23 & 0.524, 0.474, 0.498 & 223K\\
CoFE~\cite{DBLP:conf/eccv/LiLQLCEC24} & ECCV24 & 0.489, 0.370, 0.405 & 223K\\
COMG~\cite{DBLP:conf/wacv/GuLL024} & WACV24 & 0.424, 0.291, 0.345 & 223K\\
MPO~\cite{DBLP:conf/aaai/Xiao0L0B25} & AAAI25 & 0.436, 0.376, 0.353 & 223K\\
MAN~\cite{DBLP:conf/aaai/ShenPLT24} & AAAI24 & 0.411, 0.398, 0.389 & 223K\\
Med-LLM \cite{DBLP:conf/mm/LiuLZCC024} & MM24 & 0.412, 0.373, 0.395 & 223K\\
B-LLM~\cite{DBLP:conf/aaai/LiuTCS024} & AAAI24 & 0.465, 0.482, 0.473 & 223K\\
EKAGen \cite{DBLP:conf/cvpr/BuLYD24} & CVPR24 & 0.517, 0.483, 0.499 & 223K\\
MambaXray-L~\cite{DBLP:conf/cvpr/WangWLMW0025} & CVPR25 & 0.561, 0.460, 0.505 & 1.27M \\
MLRG~\cite{DBLP:conf/cvpr/LiuMK0XJM25} & CVPR25 & 0.549, 0.468, 0.505 & 240K \\
\textbf{Ours} & -- & 0.342, 0.811, 0.481 & 1K\\
% eval_qwen25_3b_grpo__gen_report__factscore__imbalanced_1000set_gptlabel__orapo_checkpoint-2750_results
\bottomrule
\end{tabular}
}
\label{tab:rrg_micro}
\end{table}

\begin{algorithm*}[!t]
\caption{The proposed OraPO with FastS Reward for Radiology Report Generation}
\label{alg:orapo}
\textbf{Input} initial policy model $\pi_{\theta}^{\text{init}}$; dataset $\mathcal{D}$ with studies $(x, y^\star)$ and prompts $p$; label set $L$; group size $K$; hyperparameters $\varepsilon,\lambda_{\mathrm{KL}},\tau,\alpha,\gamma,w_{\min},w_{\max},\beta,\xi$; outer loops $I$, inner steps $M$.\\
\textbf{Output} $\pi_\theta$
\begin{algorithmic}[1]
\State policy model $\pi_\theta \leftarrow \pi_{\theta}^{\text{init}}$
\For{$\text{iteration} = 1,\ldots,I$}
  \State reference model $\pi_{\mathrm{ref}} \leftarrow \pi_\theta$
  \For{$\text{step} = 1,\ldots,M$}
    \State Sample a minibatch $\{(x_i,p_i,y_i^\star,z_i^\star)\}^{B} \subset \mathcal{D}$
    \State Update the behaviour policy $\pi_{\theta_{\text{old}}} \leftarrow \pi_\theta$
      \State Sample $K$ reports $\{\hat{y}_{i,j}\}_{j=1}^{K} \sim \pi_{\theta_{\text{old}}}(\cdot \mid x_i,p_i)$ for each input radiology image $x_i$ in the batch
      \State {Compute FactS reward}: $\{r(x_i,\hat y_{i,j})\}_{j=1}^{K}$ 
    for each sampled report $\hat{y}_{i,j}$;
    \For{$\text{OraPO iterations} = 1,\ldots,\mu$}
      \State {Compute GRPO objective}: $L_{\mathrm{GRPO}}(x_i,p_i)$;
      \State {Compute DPO objective}: $L_{\mathrm{DPO}}(x_i,p_i)$;
      \State {Compute OraPO objective}: $\mathcal{L}_{\mathrm{OraPO}}$ by mixing $L_{\mathrm{DPO}}(x_i,p_i)$ and $L_{\mathrm{GRPO}}(x_i,p_i)$;
    \State Update the policy model $\pi_\theta$ by minimizing $L_{\mathrm{OraPO}}(x_i,p_i)$.
    \EndFor
  \EndFor
\EndFor
\end{algorithmic}
\end{algorithm*}

\section{More experimental results and analysis on MIMIC-CXR} \label{sec.mimiccxr_micro}

Table \ref{tab:rrg_micro} complements our main macro-averaged results by reporting micro-averaged Precision, Recall, and F1 on MIMIC-CXR. Macro averaging, used in the main paper, treats each disease label equally and highlights performance on rare findings. Micro averaging instead weights labels by their frequency, answering a different question: across all positive label instances in the dataset, how many do we miss. This view is especially focusing on common conditions which dominate the label distribution.

Under this micro-averaging metric, OraPO remains highly competitive while using 2--3 orders of magnitude less data. With only 1 K training reports, OraPO attains a micro recall of 0.811 and micro F1 of 0.481. The best baseline recall is EKAGen at 0.483, so OraPO improves micro recall by 67.9\% relative to this strong model. Clinically, such high recall means that, across all truly abnormal cases, our method misses far fewer findings, which is critical because false negatives on common diseases are usually more harmful than extra false positives.

Although our method achieves the highest recall, its micro F1 remains close to that of the strongest fully supervised approaches. The best baseline F1 is 0.505 (MambaXray-L and MLRG), whereas OraPO reaches 0.481, only a gap of 0.024. This trade-off is achieved with extreme data efficiency: compared with methods trained on the full 223K MIMIC-CXR split, OraPO uses 99.6\% fewer labeled reports, and compared with MambaXray-L’s 1.27M training pairs, about 99.9\% fewer. Overall, the micro results show that OraPO delivers clinically preferred high recall with competitive F1, while training with only 1K samples.

\section{Qualitative Study} \label{sec.qualitative}

In Figure \ref{fig:visual_report}, we present three examples that demonstrate how OraPO with FactS reward generates clinically accurate reports while maintaining factual grounding through atomic fact extraction and entailment checking.

\textbf{Example 1} showcases ideal performance on a challenging case with cardiomegaly and edema. The model produces a clinically coherent narrative that captures the typical imaging pattern of cardiac-related pulmonary edema. Generated facts such as ``cardiac silhouette is enlarged" and ``diffuse interstitial opacities seen centrally" directly map to the ground-truth labels, with no false positives. 

\textbf{Example 2} highlights OraPO's handling of complex cases with five ground-truth labels. The generated report correctly identifies all target pathologies, including the challenging bilateral nodular pulmonary lesions. While the model predicts pleural effusion absent from the ground-truth labels, the generated facts (``pleural effusions present" with ``decreased interval changes") suggest a false positive predicted by the model. 

\textbf{Example 3} illustrates robust multi-pathology detection spanning four conditions: lung opacity, consolidation, pneumonia, and pleural effusion. The model generates anatomically precise descriptions (``left perihilar opacity with patchy distribution") that provide localization detail often absent in baseline systems. All ground-truth labels receive explicit supporting facts, demonstrating the FactS reward's effectiveness in maintaining high recall across co-occurring pathologies. The model correctly attributes findings to specific lung regions, which aids radiologist review.

Across these examples, OraPO consistently produces factually grounded reports where each diagnostic statement traces back to verifiable atomic facts that entail specific labels. This stands in contrast to fluency-optimized baselines that generate plausible prose without guaranteed factual alignment. The FactS reward's sentence-level supervision enables the model to maintain clinical accuracy even when handling rare pathologies or multi-label complexity, directly addressing the recall-precision trade-off critical in clinical deployment.

Here’s a concise qualitative analysis paragraph you can place right below Fig. \ref{fig:visual_report}:

\subsection{Qualitative NLG Metric Analysis}
\begin{figure*}[h]
\centering
\resizebox{\textwidth}{!}{
\begin{tabular}{lp{4cm}p{3cm}p{8cm}p{6cm}p{2cm}}
\hline \toprule [0.5 pt] 
\multicolumn{1}{c}{\textbf{Image}} & \multicolumn{1}{c}{\textbf{GT Reports}} & \multicolumn{1}{c}{\textbf{GT Labels}}  & \multicolumn{1}{c}{\textbf{Our Generated Reports}} &  \multicolumn{1}{c}{\textbf{Predicted Labels and Supporting Facts}} &  \multicolumn{1}{c}{\textbf{ROUGE-L F1}}  \\

\hline \toprule [0.5 pt] 

\begin{minipage}[b]{0.4\columnwidth}\vspace{5pt}\centering\raisebox{-.9\height}{\includegraphics[width=\linewidth]{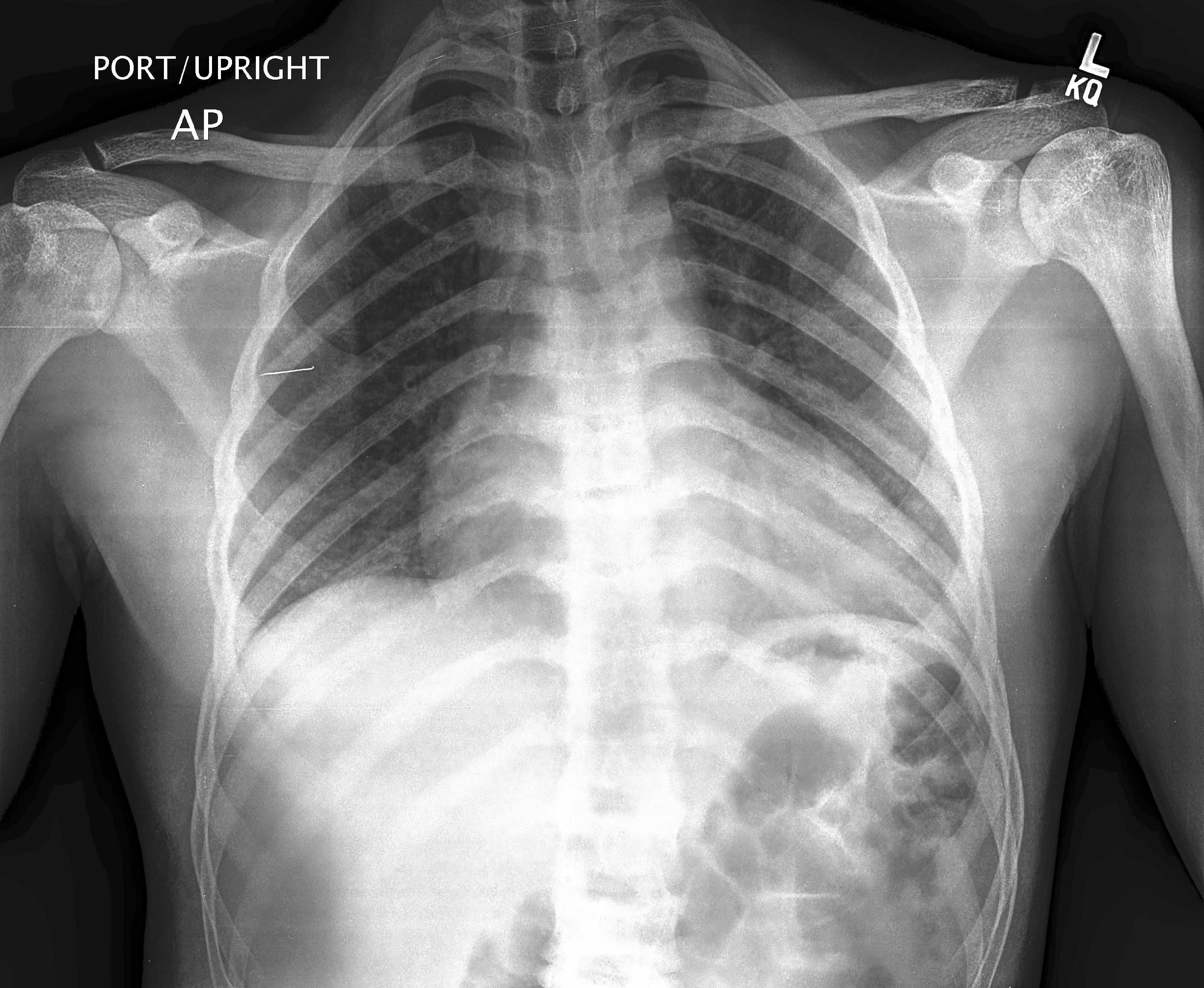}}\end{minipage}

% train/patient15962/study1/view1_frontal.jpg
&
Single frontal view of the chest demonstrates indistinct  pulmonary vasculature bilaterally, likely representing mild \matching{pulmonary edema}. \matching{Mild cardiomegaly is seen}, unchanged from prior  exam. No soft tissue or bony abnormalities are seen.
&
\matching{Cardiomegaly},

\matching{Edema}

&
Ribs: The thoracic rib contour is smooth and complete. There is mild opacity seen in the mid to left hemithorax.  Pectoral Girdle: Both clavicles are seen in their normal positions.  Diaphragm: The left diaphragm appears increased in curvature without significant deviation.  Aortic Arch/Trunk: The aortic arch appears well defined.  Cardiomediastinum: There are diffuse interstitial opacities seen centrally on the image, consistent with  \matching{pulmonary edema}. The right border of the heart is poorly defined and the \matching{cardiac silhouette is enlarged}. The pulmonary vascularity is increased centrally but diffuse through the lung fields and the cardiac silhouette appears to be expanding to the right of the sternum.  Pulmonary Vasculature: The pulmonary vasculature appears increased centrally. Lung Opacity: There is diffuse sub-pleural opacity centrally, with extension towards the left hemithorax. Increased pulmonary vasculature seen centrally, consistent with pulmonary edema. No other areas of consolidation seen. This appearance is consistent with pulmonary edema. There are small sub-pleural opacities seen centrally.

&

- \matching{Cardiomegaly}: 'Cardiac silhouette is enlarged' and 'Cardiac silhouette appears to be expanding to the right of the sternum' both indicate cardiomegaly.

- \matching{Edema}: 'Diffuse interstitial opacities seen centrally', 'Pulmonary vascularity is increased centrally', 'Pulmonary vascularity is diffuse through the lung fields', and 'Appearance is consistent with pulmonary edema' all indicate edema."

& 0.102

 \\

\begin{minipage}[b]{0.4\columnwidth}\vspace{5pt}\centering\raisebox{-.9\height}{\includegraphics[width=\linewidth]{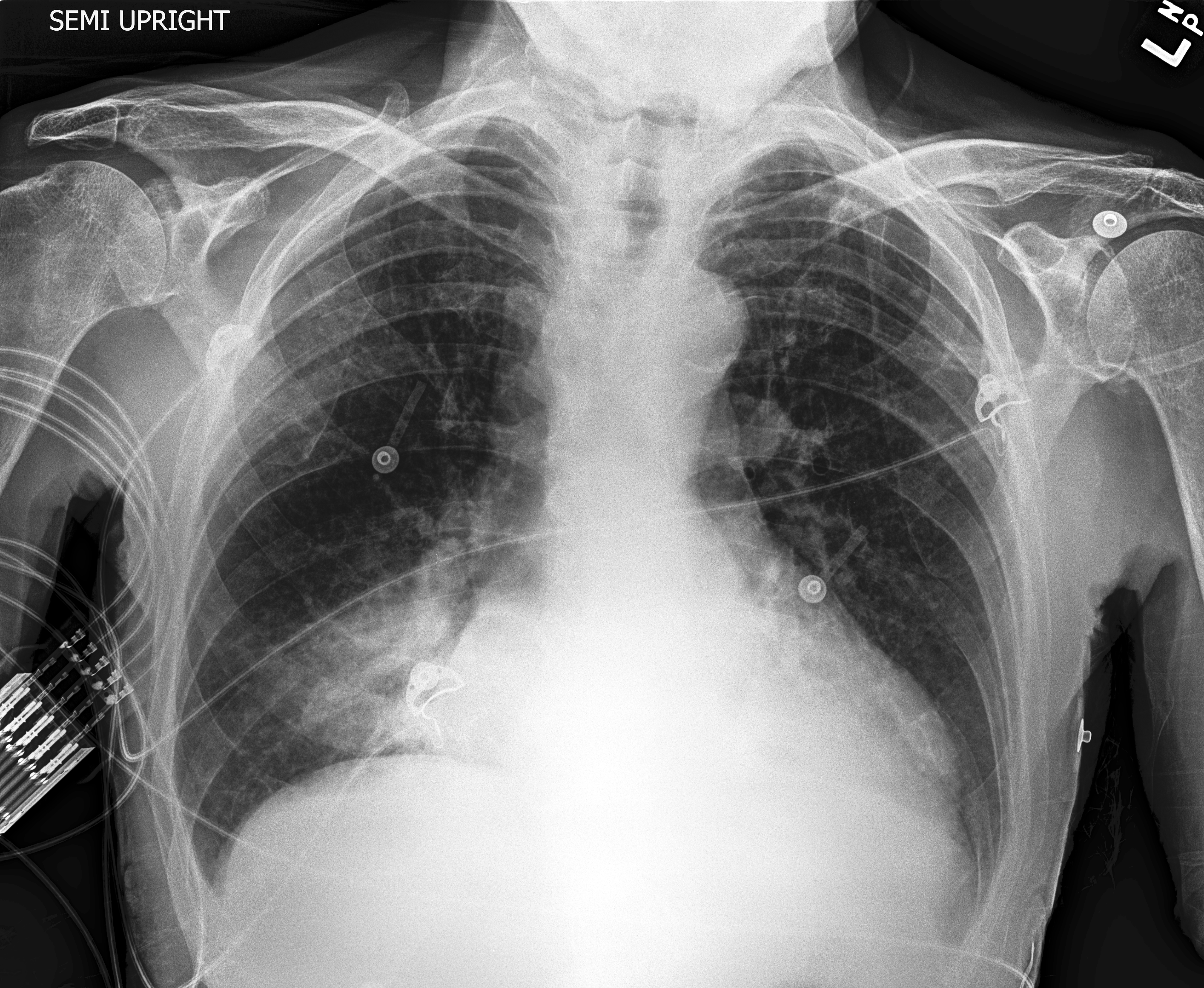}}\end{minipage}

% train/patient07755/study1/view1_frontal.jpg,
&
Within the right middle lobe there is a \matching{large  consolidation}, \matching{consistent with pneumonia}, which obscures the right  heart border.  A small left upper lobe calcified granuloma is  present.  There is mild calcification of the aortic arch.  \mismatching{No pleural  effusions}.  \matching{Mild cardiomegaly}.  No bony abnormalities are appreciated.
&
\matching{Cardiomegaly}, 

\matching{Lung Opacity},

\matching{Lung Lesion}, 

\matching{Consolidation}, 

\matching{Pneumonia}

&

\matching{Cardiomediastinal silhouette: mildly increased} in  size; cardiac contour is flattened Nodular \matching{pulmonary lesions present} bilaterally - no growth or interval increase \mismatching{Pleural effusions present} - decreased interval changes Pneumonic opacities present,  right mid and left lobe, bilateral,  but limited in extent Right diaphragmatic elevation
&

- \matching{Cardiomegaly}: 'Cardiomediastinal silhouette is mildly increased in size' directly supports presence of cardiomegaly. 

- \matching{Lung Opacity}: Facts 'Pneumonic opacities present in right mid lobe', 'Pneumonic opacities present in left lobe', and 'Nodular pulmonary lesions present bilaterally' all describe opacities.

- \matching{Lung Lesion}: 'Nodular pulmonary lesions present bilaterally' directly supports presence of lung lesions.

- \matching{Consolidation}: Facts 'Pneumonic opacities present in right mid lobe', 'Pneumonic opacities present in left lobe', and 'Pneumonic opacities are bilateral' support presence of consolidation.

- \matching{Pneumonia}: Facts 'Pneumonic opacities present in right mid lobe', 'Pneumonic opacities present in left lobe', and 'Pneumonic opacities are bilateral' support presence of pneumonia.

- \mismatching{Pleural Effusion}: 'Pleural effusions present' directly supports presence of pleural effusion.

&
0.114

 \\

% \hline \toprule [0.5 pt] 

% &
% The cardiomediastinal silhouette is unremarkable.  There is a focal airspace opacity in the right upper lobe which measures approximately 3.5 x 2.4 cm in size, and appears more confluent and discrete than on the prior chest CT from 6-29-2006.  The left lung is clear.  No pleural effusions.  The visualized osseous structures are unremarkable.", 

% &

% \matching{Lung Opacity}, 

% \matching{Lung Lesion}, 

% \matching{Consolidation}

% &

% The right lung is over-inflated and a small infiltrate is apparent inferiourly and superiourly behind the hilum. There is a patch of calcification to the right in the soft tissue plane.

% &

% - \matching{Lung Opacity}: Facts describing small infiltrates behind the right hilum indicate lung opacities. 

% - \matching{Lung Lesion}: Facts describing small infiltrates behind the right hilum indicate a lung lesion. 

% - \matching{Consolidation}: Facts describing small infiltrates behind the right hilum support the presence of consolidation.", "- Other Findings: Facts describe over-inflation and soft tissue calcification, which are not covered by other GT labels, so 'Other Findings' is present.

%  \\

\hline \toprule [0.5 pt] 

\begin{minipage}[b]{0.4\columnwidth}\vspace{5pt}\centering\raisebox{-.9\height}{\includegraphics[width=\linewidth]{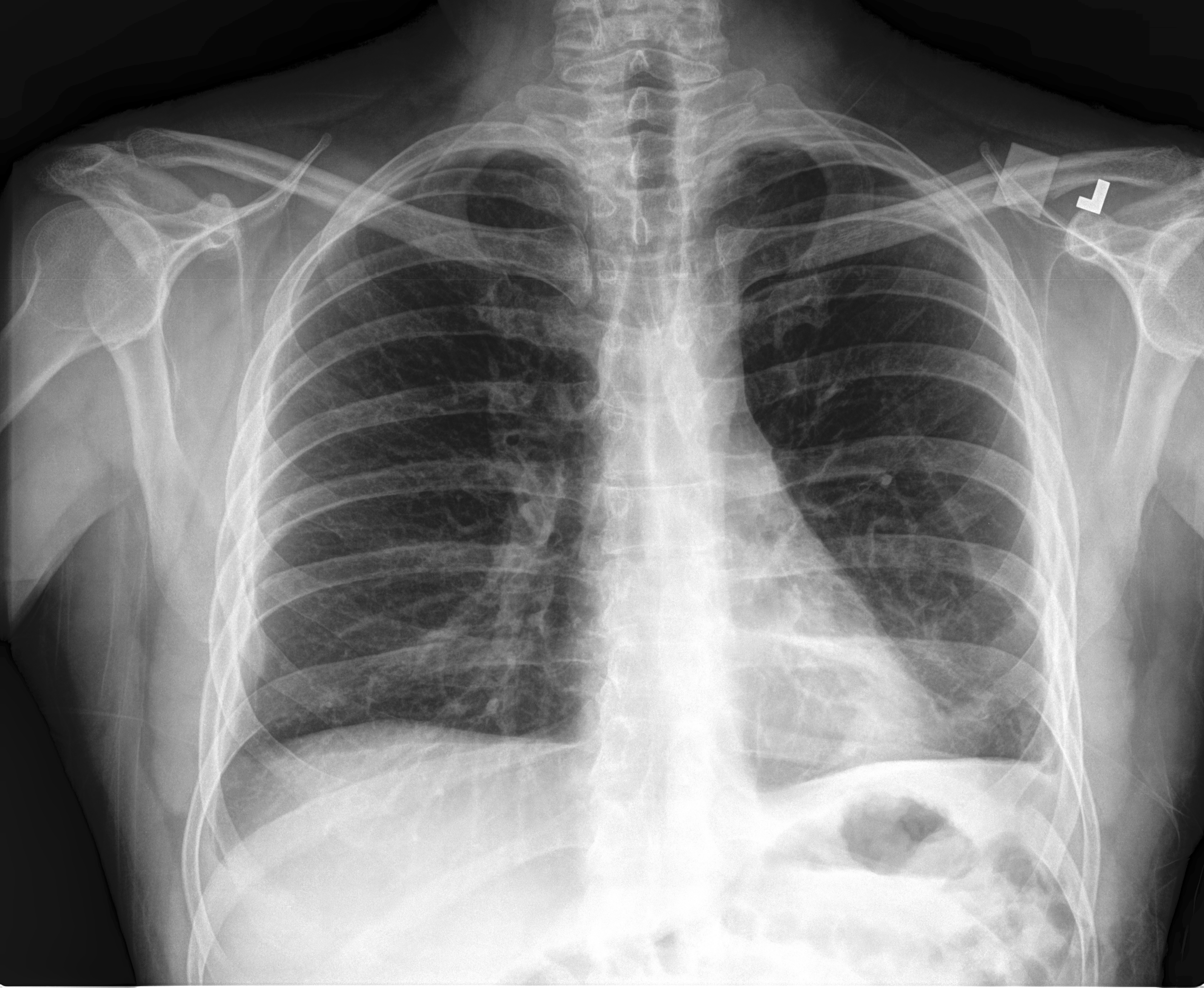}}\end{minipage}

% train/patient34576/study1/view1_frontal.jpg

&
Initial chest radiograph at 1130 hours demonstrates left lower lobe focal \matching{air space opacity which could represent pneumonia}. The remainder of the lungs are clear. The cardiomediastinal silhouette is normal. There is blunting of the left costophrenic angle which could represent \matching{a small pleural effusion.} Follow-up radiograph 1530 hours demonstrates no interval change.

&
\matching{Lung Opacity}, 

\matching{Consolidation}, 

\matching{Pneumonia}, 

\matching{Pleural Effusion} 
&
Single ventricular image. Cardiac silhouette is no  grossly enlarged (LVED: 55-57 mm, LVED: 55-57 mm)  and changes relative to previous films. No evidence  of marked cardiomegaly in the presence of  mild respiratory straining. \matching{Left perihilar opacity}  with patchy distribution. \matching{Small pleural effusion}  present, mild volume and interval increase noted.Cardiac silhouette interval increase noted.
&
- \matching{Lung Opacity}: Fact 'Left perihilar opacity with patchy distribution' describes an opacity, confirming existence.

- \matching{Consolidation}: Fact 'Left perihilar opacity with patchy distribution' describes an opacity consistent with consolidation.

- \matching{Pneumonia}: Fact 'Left perihilar opacity with patchy distribution' describes an opacity that could represent pneumonia.

- \matching{Pleural Effusion}: Facts 'Small pleural effusion present', 'Mild volume increase in pleural effusion', and 'Interval increase in pleural effusion' all confirm pleural effusion.

& 0.164

 \\

\hline \toprule [0.5 pt] 

\end{tabular}
}
\caption{X-ray image and its corresponding ground-truth, along with the output of our model generation report on the ChexPert Plus dataset. The mismatch sentence in the reports are highlighted using different colors.}
\label{fig:visual_report}
\end{figure*}

Although the ROUGE-L F1 values in Figure \ref{fig:visual_report} are modest, they largely \textbf{reflect surface mismatch rather than clinical errors}: OraPO is not trained to imitate the phrasing or template of GT reports. Instead of imitating the GT writing style, our method directly optimizes a fact-level reward. The model first extracts atomic clinical statements, and then checks whether these statements are entailed by the ground-truth labels. This training objective naturally prioritizes clinical correctness and verifiability, not surface-level similarity. In contrast, reference-overlap metrics such as ROUGE mainly measure wording overlap. They reward matched phrasing but do not penalize factual mistakes, missing findings, or even cross-sentence contradictions in long reports. Concretely, our FactScore-based reward treats the report as its own rationale, maps sentences to label-level predictions, and yields dense, interpretable feedback to the policy.  In the three qualitative cases, the generated statements are clinically grounded and aligned with the GT labels: Example 1 captures cardiomegaly and edema with supporting facts; Example 2 correctly identifies all target pathologies (with one flagged false-positive effusion); and Example 3 localizes a left perihilar opacity and supports consolidation/pneumonia/effusion—all consistent with the label set despite low lexical overlap with GT wording. This behavior is consistent with our quantitative results. OraPO is deliberately recall-oriented, which is clinically the safer operating point. On CheXpert Plus, it achieves state-of-the-art label-level recall and F1 while using only about 0.1\% of the usual training data, and it also attains very high micro-recall. Therefore, low n-gram overlap with the reference should \textbf{not be interpreted as low clinical faithfulness}.

\clearpage

% {
%     \small
%     \bibliographystyle{ieeenat_fullname}
%     \bibliography{main}
% }

% WARNING: do not forget to delete the supplementary pages from your submission 
% \input{sec/X_suppl}

% \end{document}

{
    \small
    \bibliographystyle{ieeenat_fullname}
    \bibliography{main}
}

% WARNING: do not forget to delete the supplementary pages from your submission 
% \input{sec/X_suppl}

\end{document}